\documentclass{article}

\usepackage[preprint]{ewrl_2023}

\usepackage[utf8]{inputenc} 
\usepackage[T1]{fontenc}    
\usepackage{hyperref}       
\hypersetup{
    colorlinks=true,
    linkcolor=blue,
    citecolor=blue,
    filecolor=blue,
    urlcolor=blue,
}
\usepackage{url}            
\usepackage{booktabs}       
\usepackage{amsfonts}       
\usepackage{nicefrac}       
\usepackage{microtype}      

\usepackage{amsthm}

\newcommand{\eqq}[1]{\begin{align}#1\end{align}}

\usepackage{cancel}

\def\G{\mathcal{G}}

\def\E{\mathbb{E}}
\def\R{\mathbb{R}}

\usepackage{nicefrac}

\def\eps{{\epsilon}}
\def\Adv{\operatorname{A}}
\def\A{\mathscr{A}}

\def\F{\mathbf{F}}
\def\R{\mathbb{R}}

\usepackage{algorithmic}
\usepackage{algorithm}
\usepackage{wrapfig}

\def\T{\mathcal{T}}

\def\P{\mathcal{P}}

\def\S{\mathscr{S}}

\def\R{\mathbb{R}}

\def\F{\mathbf{F}}

\def\F{\mathbf{F}}

\def \eps{{\epsilon}}

\def\KL{\operatorname{KL}}

\def\sigma{\lambda}
\usepackage{algorithmic}
\usepackage{algorithm}
\usepackage{wrapfig}

\def\B{{\mathscr{B}}}

\usepackage[mathscr]{eucal}

\usepackage{enumitem}

\usepackage[dvipsnames]{xcolor}

\definecolor{yellow}{rgb}{0.75,0.65,0.15}
\definecolor{sea_green}{HTML}{3CB371}
\definecolor{cornflower_blue}{HTML}{6495ED}
\definecolor{retro_orange}{HTML}{DE7D37}
\definecolor{earth_red}{HTML}{AD3956}

\definecolor{turquoise}{rgb}{0.3,0.7,0.7}

\definecolor{dark turquoise}{HTML}{01665e}
\definecolor{brown}{rgb}{0.6,0.3,0.}

\definecolor{light brown}{HTML}{bf812d}

\usepackage{microtype}
\usepackage{graphicx}
\usepackage[dvipsnames]{xcolor}

\usepackage{kantlipsum, lipsum}
\usepackage{dsfont}
\usepackage{subfigure}
\usepackage{booktabs} 

\usepackage{amsmath}
\usepackage{amssymb}
\usepackage{mathtools}
\usepackage{amsthm}
\RequirePackage{algorithm}
\RequirePackage{algorithmic}
\usepackage{amsthm}
\def\clubsuit{\ding{168}}              
\def\varheartsuit{{\color{black}\ding{170}}} 
\def\spadesuit{\ding{171} }             

\usepackage{caption}
\usepackage{graphicx}
\usepackage{tabularx}
\usepackage{mwe}
\newsavebox{\tempbox}

\usepackage{float}
\usepackage{multirow}
\usepackage{pifont}

\newtheoremstyle{theorem}
  {\topsep}
  {\topsep}
  {}
  {}
  {\itshape}
  {}
  {.5em}
  {\thmname{#1}\thmnumber{ #2}\thmnote{ (#3)}}
\theoremstyle{theorem}

\usepackage{parskip}

\newtheorem*{example}{Example}

\counterwithin{lemma}{section}
\counterwithin{definition}{section}
\counterwithin{remark}{section}
\counterwithin{observation}{section}
\counterwithin{assumption}{section}

\usepackage[font={footnotesize}, labelfont={bf,footnotesize}]{caption}
\usepackage{authblk}
\DeclareFontFamily{U}{matha}{\hyphenchar\font45}
\DeclareFontShape{U}{matha}{m}{n}{
      <5> <6> <7> <8> <9> <10> gen * matha
      <10.95> matha10 <12> <14.4> <17.28> <20.74> <24.88> matha12
      }{}
\DeclareSymbolFont{matha}{U}{matha}{m}{n}
\DeclareFontSubstitution{U}{matha}{m}{n}

\DeclareFontFamily{U}{mathx}{\hyphenchar\font45}
\DeclareFontShape{U}{mathx}{m}{n}{
      <5> <6> <7> <8> <9> <10>
      <10.95> <12> <14.4> <17.28> <20.74> <24.88>
      mathx10
      }{}
\DeclareSymbolFont{mathx}{U}{mathx}{m}{n}
\DeclareFontSubstitution{U}{mathx}{m}{n}

\DeclareMathDelimiter{\vvvert}{0}{matha}{"7E}{mathx}{"17}
\allowdisplaybreaks
\title{
Acceleration in Policy Optimization}

\makeatletter \renewcommand\AB@affilsepx{\quad  \protect\Affilfont} \makeatother

\author[\varheartsuit\spadesuit]{Veronica Chelu \thanks{
Correspondence: <\texttt{veronica.chelu@mail.mcgill.ca}>. Work executed on internship at DeepMind.}}

\author[\clubsuit]{Tom Zahavy}
\author[\clubsuit]{Arthur Guez}
\author[\varheartsuit\spadesuit\clubsuit]{Doina Precup}
\author[\clubsuit]{Sebastian Flennerhag}
\affil[\varheartsuit]{McGill University}
\affil[\spadesuit]{Mila Quebec AI Institute}
\affil[\clubsuit]{Google DeepMind}

\begin{document}

\maketitle

\begin{abstract}
  We work towards a unifying paradigm for accelerating policy optimization methods in reinforcement learning (RL) by integrating foresight in the policy improvement step via optimistic and adaptive updates.
Leveraging the connection between policy iteration and policy gradient methods, we view policy optimization algorithms as iteratively solving a sequence of surrogate objectives, local lower bounds on the original objective. We define optimism as predictive modelling of the future behavior of a policy, and adaptivity as taking immediate and anticipatory corrective actions to mitigate accumulating errors from overshooting predictions or delayed responses to change. 
We use this shared lens to jointly express other well-known algorithms, including model-based policy improvement based on forward search, and optimistic meta-learning algorithms. 
We analyze properties of this formulation, and show connections to other accelerated optimization algorithms. Then, we design an optimistic policy gradient algorithm, adaptive via meta-gradient learning, and empirically highlight several design choices pertaining to acceleration, in an illustrative task.
\end{abstract}

\section{Introduction}\label{sec:introduction}
Policy gradient (PG) methods \citep{Williams92, Sutton2000} are one of the most effective reinforcement learning (RL) algorithms \citep{impala, Schulman15, Schulman17, Abdolmaleki18, Hessel2021, Zahavy20, Flennerhag22}. These methods search for the optimal policy in a parametrized class of policies by using gradient ascent to maximize the cumulative expected reward that a policy collects when interacting with an environment.
While effective, this objective poses challenges to the analysis and understanding of PG-based optimization algorithms due to its non-concavity in the policy parametrization \citep{Agarwal19, Mei20a, Mei20b, Mei21}. 

Nevertheless, PG methods globally converge sub-linearly for smoothly parametrized softmax policy classes. This analysis relies on local linearization of the objective function in parameter space and uses small step sizes and gradient domination to control the errors introduced from the linearization \citep{Agarwal19, Mei20a, Mei21, mei2023role}. 
In contrast, policy iteration (PI) linearizes the objective w.r.t. (with respect to) the functional representation of the policy \citep{Agarwal19, BhandariandRusso_globalguarantees, bhandari21a}, and converges linearly when the surrogate objective obtained from the linearization is known and can be solved in closed form. 

Relying on the illuminating connections between PI and several instances of PG algorithms (including (inexact) natural policy gradients (NPG) and mirror ascent (MA)), recent works \citep{bhandari21a, cen, Mei22, yuan2023linear, alfano2023linear, chen22i} extended the above results and showed linear convergence of PG algorithms with large step sizes (adaptive or geometrically increasing). Other works showed that PG methods can achieve linear rates via entropy regularization. These guarantees cover some (approximately) closed  policy classes, e.g., tabular, or log-linear---cf. Table~\ref{table:all} in Appendix~\ref{apend:related_work_context}. 
More generally, in practice, each iteration of these PI-like algorithms is solved approximately, using a few gradient ascent update steps, in the space of policy parameters, which lacks guarantees due to non-concavity induced by non-linear transformations in the deep neural networks used to represent the policy \citep{Agarwal19, Abdolmaleki18, Tomar20, Vaswani2021}.   

This recent understanding about the convergence properties of policy gradient methods in RL leaves room to consider more advanced techniques. In this work, we focus on \textbf{acceleration} via \emph{optimism}---a term we borrow from online convex optimization \citep{Zinkevich2003}, and is unrelated to the exploration strategy of \emph{optimism in the face of uncertainty}.
In this context, \emph{optimism} refers to predicting future gradient directions in order to accelerate convergence (for instance, as done in Nesterov's accelerated gradients (NAG) \citep{Nesterov1983AMF, WangAbernethy2018,Wang2021}, extra-gradient (EG) methods \citep{Korpelevich1976TheEM},  mirror-prox \citep{Nemirovski04, juditsky2011solving}, optimistic MD \citep{RakhlinS13, joulani20a}, AO-FTRL \citep{rakhlin2014online, mohri2015accelerating}, etc.). 

In RL, optimistic policy iteration (OPI) \citep{bertsekas:ndp-book, Bertsekas2011, Tsitsiklis2002} considers policy updates performed based on incomplete evaluation, with a value function estimate that gradually tracks the solution of the most recent policy evaluation problem. Non-optimistic methods, on the other hand, regard the value estimation problem as a series of isolated evaluation problems and solve them by Monte Carlo or temporal difference (TD) estimation. By doing so, they ignore the potentially \emph{predictable} nature of the evaluation problems, and their solutions, along a policy's optimization path. 

In previous work, optimism has been studied in policy optimization to mitigate oscillations \citep{Wagner14, NIPS2013_2dace78f, moskovitz2023reload} as well as for accelerated optimization \citep{predictor_corrector, Hao2020}, resulting in sub-linear, yet unbiased convergence, cf. Table~\ref{table:all} in Appendix~\ref{apend:related_work_context}.

In this paper, we introduce a general policy optimization framework that allows us to describe seemingly disparate algorithms as making specific choices in how they represent, or adapt optimistic gradient predictions. 
Central to our exposition is the idea of \emph{prospective learning}, i.e. making \emph{predictions} or projections of the future behavior, performance, or state of a system, based on existing historical data (\emph{interpolation}), or extending those predictions into uncharted territory by predicting beyond data  (\emph{extrapolation}). This learning approach  explicitly emphasizes the ability to anticipate the future when a recognizable pattern exists in the sequence. 

In particular, we show that two classes of 
well-known algorithms---\emph{meta-learning algorithms} and \emph{model-based planning algorithms}---can be viewed as optimistic variants of vanilla policy optimization, and provide a theoretical argument for their empirical success. For example, STACX \citep{Zahavy20} represents an optimistic variant of Impala \citep{impala} and achieves a doubling of Impala's performance on the Atari-57 suite; similarly, adding further optimistic steps in BMG \citep{Flennerhag22} yields another doubling of the performance relative to that of STACX. In model-based RL, algorithms with extra steps of planning, e.g., the AlphaZero family of algorithms \citep{silver2016, silver2017}, with perfect simulators, also enjoy huge success in challenging domains, e.g. chess, Go, and MuZero \citep{Schrittwieser2019}, with an \emph{adaptive} model, achieves superhuman performance in challenging and visually complex domains.

\paragraph{Contributions}\quad
After some background in Sec.~\ref{sec:background_and_notation}, we define a simple template for accelerating policy optimization algorithms in Sec.~\ref{sec:op_view}. This formulation involves using \emph{proximal policy improvement methods} with \emph{optimistic auto-regressive policy update rules}, which adapt to anticipate the future policy performance.
We show this \emph{acceleration} template based on optimism \& adaptivity is a generalization of the update rule of proximal policy optimization algorithms, where the inductive bias is \emph{fixed}, and does not change with past experience. We use the introduced generalization  to show that a \emph{learned} update rule can form other inductive biases, that can accelerate convergence. 

We use the introduced formulation to highlight the commonalities among several algorithms, expressed in this formalism in Sec.~\ref{sec:op_view}, including model-based policy optimization algorithms relying on run-time forward search (e.g. \citet{silver2016, silver2017, Schrittwieser2019, Hessel2021}), and a general algorithm for \emph{optimistic policy gradients} via \emph{meta-gradient optimization} (common across the algorithmic implementations of \citet{Zahavy20, Flennerhag22}).

 Leveraging theoretical insights from Sec.~\ref{sec:op_view}, in Sec.~\ref{sec:optimistic_policy_gradients}, we introduce an optimistic policy gradient algorithm that is adaptive via meta-gradient learning. In Sec.~\ref{sec:empirical_analysis}, we use an illustrative task to test several theoretical predictions empirically. First, we tease apart the role of optimism in forward search algorithms. Second, we analyze the properties of the optimistic algorithm we introduced in Sec.~\ref{sec:optimistic_policy_gradients}.

 Using acceleration for functional policy gradients is under-explored, and we hope this unifying template can be used to design other accelerated policy optimization algorithms, or guide the investigation into other collective properties of these methods.

\section{Preliminaries \& notation}\label{sec:background_and_notation}
\paragraph{Notation} Throughout the manuscript, we use $\doteq$ to distinguish a definition from standard equivalence, the shorthand notation $\nabla_{x} f(x_t) \doteq \nabla_{x} f(x)|_{x=x_t}$, $\langle\cdot, \cdot\rangle$ denotes a dot product between the arguments. The notation $\lceil\cdot\rfloor$ indicates that gradients are not backpropagated through the argument.

\subsection{Markov Decision Processes}{
We consider a standard reinforcement learning (RL) setting described by means of  a discrete-time infinite-horizon discounted Markov decision process (MDP) \citep{puterman} $\mathscr{M}\doteq \{ \S, \A, r, P, \gamma, \rho\}$, with state space $\S$ and action space $\A$, discount factor $\gamma\in [0, 1)$, with initial states sampled under the initial distribution $\rho$, assumed to be exploratory $\rho(s) > 0, \forall s \in \S$.

The agent follows an online learning protocol: at timestep $t \geq 0$, the agent is in state $S_t \in \S$, takes action $A_t \in \A$, given a policy $\pi_t(\cdot|s_t)$---the distribution over actions for each state $\pi : \S\to \Delta_{\A}$, with $\Delta_{\A}$---the action simplex---the space of probability distributions defined on  $\A$. It then receives a reward $R_t \sim r(S_t, A_t)$, sampled from the reward function $r:\S\times\A\to[0,R_{\max}]$,  and transitions to a next state $S_{t+1} \sim P(\cdot|S_t, A_t)$, sampled under the transition probabilities or dynamics $P$. 
Let $d_{\pi}(s)$ be a measure over states, representing the discounted visitation distribution (or discounted fraction of time the system spends in a state $s$) $d_{\pi}(s)= (1-\gamma) \sum_{t=0}^{\infty} \gamma^t \operatorname{Pr}(S_t=s|S_0\sim \rho, A_k \sim \pi(\cdot|S_k),\forall k\leq t)$, with $\operatorname{Pr}(S_t=s|S_0\dots)$ the probability of transitioning to a state at timestep $t$ given policy $\pi$.

The RL problem consists in finding a policy $\pi$ maximizing the discounted return
\eqq{
J(\pi)\doteq \E_{S\sim\rho}[V_\pi(S)] = (1-\gamma)\mathbb{E}_{\pi, \rho}\left[\textstyle\sum_{t\geq 0}\gamma^t R_{t+1}\right] &&\text{(\emph{the policy performance objective})} \label{eq:pg_loss}
}
where $V_\pi \in\R^{|\S|}$ is the value function, and $Q_\pi\in \R^{|\S|\times|\A|}$ the action-value function of a policy $\pi\in \Pi= \{ \pi \in \mathbb{R}^{|\S|\times |\A|}_+ | \sum_{a\in\A} \pi(s,a)= 1, \forall s \in\S\}$, s.t. (such that) $Q_\pi(s,a) \doteq \E_\pi \left[\sum_{t=0}^\infty \gamma^t R_t|S_0= s, A_0=a\right]$, and $V_\pi(s) \doteq \E_\pi\left[Q(s,A)\right]$.
Let $\T_\pi: \R^{|\S|}\to \R^{|\S|}$ be the Bellman evaluation operator, and $\T: \R^{|\S|} \to \R^{|\S|}$ the Bellman optimality operator, s.t. $(\T_\pi V) (s)\doteq r(s, \pi(s))+ \gamma \sum_{s^\prime \in \S} P(s^\prime|s, \pi(s)) V(s^\prime)$, and $(\T V)(s) \doteq \max_{a\in \A} r(s, a)+ \gamma \sum_{s^\prime \in \S}P(s^\prime|s,a) V(s^\prime) = \max_{\pi\in\Pi} (\T_\pi V)(s)$, with Q-function (abbr. Q-fn) analogs.
}

\subsection{Policy Optimization Algorithms}{
The classic \emph{policy iteration (PI)} algorithm repeats consecutive stages of (i) one-step greedy policy improvement w.r.t. a value function estimate 
$\pi_{t+1} \in \G(V_{\pi_t}) \doteq \{\pi: \T_\pi V_{\pi_t}= \T V_{\pi_t}\}$, with ${\G}$ the greedy set of $V_{\pi_t}$,
followed by (ii) evaluation of the value function w.r.t. the greedy policy $V_{\pi_{t+1}}= \lim_{m\to \infty} \T^{m}_{\pi_{t+1}} V_{\pi_t}$. 
Approximations of either steps lead to \emph{approximate PI (API)} \citep{scherrer15a}.
Relaxing the greedification leads to \emph{soft PI} \citep{cpi}
$\pi_{t+1} \doteq (1-\alpha) \pi_t + \alpha\pi^+_{t+1}$, with
$\pi_{t+1}^+ 
\doteq \arg\max_{\pi\in\Pi} \langle  {Q}_{\pi_t}, \pi\rangle $, 
for $\alpha\in(0,1]$, a step size.
\emph{Optimistic PI (OPI)} \citep{bertsekas:ndp-book} relaxes the evaluation step instead to $Q_{t+1} \doteq  Q_t - {\lambda} [Q_t - \T Q_t ]$. Others \citep{Smirnova2020, Kavosh} have extended these results to deep RL and or alleviated assumptions.

More commonly used in practice are \emph{policy gradient} algorithms. These methods search over policies using surrogate objectives $\ell_t(\pi)$ that are local linearizations of the performance $\ell_t(\pi) \doteq  J(\pi_t) + \langle \pi, \nabla_{\pi} J(\pi_t) \rangle - \nicefrac{1}{2\alpha}\|\pi - \pi_t\|_{\Omega}^2$, rely on the true gradient ascent direction of the previous policy in the sequence $\nabla_{\pi} J(\pi_t)$, and lower bound the policy performance \citep{Agarwal19, analyUpdate2021, Vaswani2021} when $J (\pi)$ is $\frac{1}{\alpha}\Omega$-relatively convex w.r.t. the policy $\pi$ \citep{lu2017relativelysmooth, Johnson2020}, which holds when $\alpha$ is sufficiently conservative.
As $\alpha \to \infty$ (the regularization term tends to zero), $\pi_{t+1} = \arg\max_{\pi\in\Pi} \ell_t(\pi)$
converges to the solution of $\ell_t$, which is exactly the policy iteration update. For intermediate values of
$\alpha$, the \emph{projected gradient ascent} update decouples across states and takes the following form for a direct policy parametrization: $\pi_{t+1} \doteq \P_{\Pi}(\pi_{t} + \alpha  Q_{\pi_t})$, with $\P_{\Pi}$ a projection operator.

Generally, the methods employed in practice extend the policy search  to parameterized policy classes with softmax transforms $\Pi_{\Theta} \doteq \big\{ \pi_\theta  \big|  \pi_\theta(s,a) =\nicefrac{\exp z_\theta(s,a)}{\sum_a\exp z_\theta(s,a)}\forall s\in\S, a\in\A, \theta\in \Theta \subset \mathbb{R}^m\big\}$, with $z_\theta$ a differentiable function, either tabular  $z_\theta(s,a) \doteq \theta_{s,a}$, log-linear $z_\theta(s,a) \doteq \phi(s,a)^\top \theta$, with $\phi$ a feature representation, or neural parametrizations ($z_\theta$-a neural network) \citep{Agarwal19}. These methods search over the parameter vector $\theta$ of a policy $\pi_\theta \in \Pi_{\Theta}$. \emph{Actor-critic} methods approximate the gradient direction with a parametrized critic ${Q}_{w_t} \approx {Q}_{\pi_t}$, with parameters $w \in \mathcal{W} \subset \mathbb{R}^{m}$, yielding 
$\theta_{t+1} \doteq \arg\max_{\theta \in \Theta} \ell(\pi_{\theta}, Q_{w_t})$, with the surrogate objective $\ell(\pi_{\theta}, Q_{w_t}) \doteq \langle \pi_{\theta}, \hat{g}_t\rangle-1/\alpha \KL_{[d_{\pi_{\theta_t}}]}(\pi_\theta, \pi_{\theta_t})$, where $\hat{g}_t \doteq d_{\pi_{\theta_t}}^\top {Q}_{w_t}$, and we denoted $\KL_{[d]}(\pi, \pi^\prime) \doteq \sum_{s} d(s) \sum_a \pi(a|s) (\log\pi(a|s) - \log \pi^\prime(a|s))$ the weighted KL-divergence. The projected gradient ascent version of this update uses the KL-divergence---the projection associated with the softmax transform, $\pi_{t+1} \doteq \KL(\pi, \nicefrac{\exp{z_{t+\nicefrac{1}{2}}}}{\sum_{\A} \exp{z_{t+\nicefrac{1}{2}}}})$ with $z_{t+\nicefrac{1}{2}}\! =\! \log\pi_t + \alpha  Q_{\pi_t}$ a target-based update. 
}

\paragraph{Acceleration}
When the effective horizon $\gamma$ is large, close to $1$ the number of iteration before convergence of policy or value iteration, scales on the order $\mathcal{O}\left(\nicefrac{1}{1-\gamma}\right)$. Each iteration is expensive in the number of samples. One direction to accelerate is designing algorithms convergent in a smaller number of iterations, resulting in significant empirical speedups.
\textbf{Anderson acceleration} 
\cite{Anderson65} is an iterative algorithm that combines information from previous iterations to update the current guess, and allows speeding up the computation of fixed points. Anderson Acceleration has been described for value iteration in \cite{Geist2018}, extensions to Momentum Value Iteration and Nesterov's Accelerated gradient in \cite{goyal2021firstorder}, and to action-value (Q) functions in \cite{Vieillard2019}. In the following, we present a policy optimization algorithm with a similar interpretation.

\paragraph{Model-based policy optimization (MBPO)}{
MBPO algorithms based on Tree Search \citep{Coulom2006EfficientSA, silver2016, silver2017, Hallak21, Rosenberg22, dalal2023softtreemax} rely on approximate online versions of multi-step greedy improvement implemented via Monte Carlo Tree Search (MCTS) \citep{BrownePWLCRTPSC12}. These algorithms replace the one-step greedy policy in the improvement stage of PI with a multi-step greedy policy.
Cf. \citet{Grill2020}, relaxing the hard greedification, and adding approximations over parametric policy classes, forward search algorithms at scale, can be written as the solution to a regularized optimal control problem, 
by replacing the gradient estimate in the regularized policy improvement objectives $\ell(\pi)$ of actor-critic algorithms with update $U_{t}$ resulting from approximate lookahead search using a model or simulator $(\hat{r}, \hat{P})$ \citep{silver2016, silver2017,Schrittwieser2019} up to some horizon $h$, using a tree search policy $\pi_b \doteq \pi_{\theta_t}$:
$
\theta_{t+1} \doteq \arg\max_{\theta \in \Theta} \langle \pi, U_{t}\rangle_{d_{\pi_t}}-1/\alpha\KL_{[d_{\pi_t}]}(\pi, \pi_{t})$, where
$ U_{t} \doteq \hat{r}^h(s,a) + \gamma \sum_{s^\prime} \hat{P}^h(s^\prime|s,a)  \sum_{a^\prime} {\pi_b}(a^\prime|s^\prime)Q_{w_t}(s^\prime,a^\prime)$.
}
\paragraph{Meta-gradient policy optimization (MGPO)}{ In MGPO \citep{XuHS18, Zahavy20, Flennerhag22} the policy improvement step uses a parametrized recursive algorithm $
\pi_{\theta_{t+1}} = \varphi(\eta_t, \pi_{\theta_t})$ with $\eta\in\R^n$ the algorithm's (meta-)parameters.
For computational tractability, we generally apply inductive biases to limit the functional class of algorithms the meta-learner searches over, e.g., to gradient ascent (GA) parameter updates
$\theta_{t+1} = \theta_t + y_{\eta_t}$.
The meta-parameters $\eta$ can represent, e.g., inializations \citep{FinnAL17}, losses \citep{meta_critic, meta_return, Kirsch2019, Houthooft2018, meta_loss, Xu2020}, internal dynamics \citep{rl2}, exploration strategies \citep{meta_exploration, Flennerhag22}, hyperparameters \citep{Veeriah2019, XuHS18, Zahavy20}, and intrinsic rewards \citep{meta_rewards}.
The meta-learner's objective is to adapt the parametrized optimization algorithm based on the learner's post-update performance $J(\pi_{\theta_{t+1}})$---unknown in RL, and replaced with a surrogate objective $\ell(\pi_{\theta_{t+1}})$.
\citet{Zahavy20} uses a linear model,
 whereas \citet{Flennerhag22} a quasi-Newton method \citep{Nocedal2006, Martens14} by means of a trust region with a hard cut-off after $h$ parameter updates. 
}

\section{Acceleration in Policy Optimization}\label{sec:op_view}

We introduce a simple template for accelerated policy optimization algorithms, and analyze its properties for finite state and action MDPs, tabular parametrization, direct and softmax policy classes. Thereafter, we describe a practical and scalable algorithm, adaptive via meta-gradient learning. 

\subsection{A general template for acceleration}
Consider finite state and action MDPs, and a tabular policy parametrization. The following policy classes will
cause policy gradient updates to decouple across states since $\Pi \equiv \Delta_{\A}\times\dots\Delta_{\A}$ 
---the $|\S|$-fold product of
the probability simplex: (i) the \emph{direct policy representation} using a policy class consisting of all stochastic policies $\Pi = \{ \pi \in \mathbb{R}^{|\S|\times |\A|}_+ \big| \sum_{a\in\A} \pi(s,a) = 1, \forall s \in\S\}$, and (ii) the \emph{softmax policy representation} $\Pi \doteq \big\{\pi \big| \pi(s,a) = \nicefrac{\exp z(s,a)}{\sum_a\exp z(s,a)},\forall s\in\S, a\in\A \big\}$, with $z$ a dual target, the logits of a policy before normalizing them to probability distributions. Let $\Omega : \R^{|\S| \times |\A|}  \to \R^{|\S|}$, be a mapping function s.t. $z = (\nabla \Omega)^{-1}(\pi)$. For (i) the direct parametrization, we have $(\nabla \Omega)^{-1}$ the identity mapping, and $z = \pi$. For (ii) the softmax transform, we have $z = (\nabla \Omega)^{-1}(\pi)$ the logarithm function, and $\nabla \Omega(z)$ the exponential function.
A new policy is obtained by projecting $\nabla \Omega(z)$ onto the constraint set induced by the probability simplex $\Pi$, using a projection operator $\pi\doteq \P_{\Pi} \nabla \Omega(z)$. Let $\{g_t\}_{t\geq 0}$ be (functional) policy gradients, $g \doteq \nabla_{\pi} J(\pi)$, and ${\hat{g}}_t \approx g_t$ approximations, e.g. stochastic gradients, or the outputs of models or simulators. 
Let $\{u_t\}_{t\geq 0}$ be a sequence of (functional) policy updates, described momentarily.

\paragraph{Base algorithm} 
Iterative methods decompose the original multi-iteration objective in Eq.~\ref{eq:pg_loss} into single-iteration surrogate objectives $\{\ell_t(\pi)\}_{t\geq0}$, which correspond to finding a maximal policy improvement policy $\pi_{t+1}$ for a single iteration $\pi_{t+1} =\arg\max_{\pi\in\Pi}\ell_t(\pi)$ and following $\pi_t$ thereafter. 
We consider first-order surrogate objectives 
\eqq{
\pi_{t+1} \!\doteq\! \arg\max_{\pi\in\Pi} \ell_t(\pi, u_t)\qquad \ell_t(\pi, u) \!\doteq\! \langle \pi, u\rangle \!-\! \nicefrac{1}{\alpha}\|\pi \!-\! \pi_t\|^2_{\Omega} &&\text{\emph{(local surrogate objective)}}\! \label{eq:local surrogate objective} 
}
with $\alpha$ a step size set to guarantee relative convexity of $\ell$ w.r.t. $\pi$ \citep{lu2017relativelysmooth, Johnson2020}, and $\|\cdot\|_{\Omega}$ the policy distance measured in the norm induced by the policy transform $\Omega$ (Euclidean norm for the direct parametrization, and KL-divergence $\KL(\cdot, \cdot)$ for the softmax parametrization, cf. \cite{Agarwal19}). At optimality, we obtain projected gradient ascent (cf.\citet{Bubeck15, bhandari21a, Vaswani2021}) 
\eqq{
{\pi}_{t+1} &\doteq \P_{\Delta^{|\A|}} \left(\nabla \Omega(z_{t+\nicefrac{1}{2}})\right)\qquad z_{t+\nicefrac{1}{2}} = z_{t} + \alpha {u}_{t} \label{eq:optimistic_proj} &&\text{\emph{(policy improvement)}} 
}
 with $z_t \doteq (\nabla \Omega)^{-1}(\pi_t)$, and $\P_{\Pi}$ the projection operator associated with the policy class (Euclidean for the direct parametrization, and the KL-divergence for the softmax parametrization, cf. \cite{Agarwal19, bhandari21a}). It is known that for the softmax parametrization the closed-form update is the natural policy gradient update $\pi_{t+1} \propto \pi_t \exp (\alpha {u}_{t})$.
\begin{wrapfigure}{r}{0.47\textwidth} 
\begin{minipage}{0.47\textwidth} 
\vspace{-5pt}
\begin{algorithm}[H]
\caption{{Accelerated Policy Gradients (generic)}}
\label{alg:acc_pi}
\begin{algorithmic}
{\footnotesize
  \FOR{$t = 1,2 \dots T$}
         \STATE 	$\triangleright$\emph{policy evaluation}: estimate $\hat{g}_t \approx g_t$ 
         \STATE 	$\triangleright$\emph{acceleration}: update $u_t$ with momentum Eq.~\ref{eq:momentum}, optimism+lookahead Eq.~\ref{eq:optimism}, or extra-gradients Eq.~\ref{eq:extragrads}
        \STATE 	$\triangleright$\emph{policy improvement}: update $\pi_{t+1}$ with Eq.~\ref{eq:optimistic_proj}
    \ENDFOR
    }
\end{algorithmic}
\end{algorithm}
\end{minipage}
\vspace{-10pt}
\end{wrapfigure}
\paragraph{Acceleration}
If the update rule ${u}_t$ returns just an estimation of the standard gradient ${u}_t \doteq \hat{g}_t$, with $\hat{g}_t \approx g_t$,
then the algorithm reduces to the inexact NPG (mirror ascent/proximal update) ${\pi}_{t+1} \doteq \P_{\Delta^{|\A|}} (\nabla \Omega)^{-1}\left(z_t + \alpha \hat{g}_t \right)$. The inductive bias is fixed and does not change with past 
experience, and acceleration is not possible. 
If the update rule is auto-regressive, the inductive bias formed is similar to the canonical \textbf{momentum} algorithm---Polyak’s Heavy-Ball method \citep{POLYAK19641},
\eqq{
 u_{t} \!\doteq\! \mu u_{t-1} \!+\!  \beta \hat{g}_t \!\implies\!z_{t+\nicefrac{1}{2}} \!=\! z_t \!+ \!\mu(z_{t-\nicefrac{1}{2}} \!- \!z_{t-1}) \!+\! \alpha \beta\hat{g}_t  &&\text{\emph{(momentum/Heavy-Ball)}\!}\! \label{eq:momentum}
}
with $\beta$ a step-size, and $\mu$ the momentum decay value. Because Heavy Ball carries momentum from past updates, it can encode a model of the learning dynamics that leads to faster convergence.

\paragraph{Optimism}
A typical form of {optimism} is to predict the next gradient in the sequence $\hat{g}_{t+1} \approx g_{t+1}$, while  simultaneously subtracting the previous prediction $\hat{g}_t$, thereby adding a tangent at each iteration
 \eqq{
u_t  \!\doteq\! \beta \hat{g}_{t+
1}\! +\! \mu[{u}_{t-1} \!- \! \beta \hat{g}_{t}] \!\implies\! z_{t+\nicefrac{1}{2}} \!=\! z_t \!+\! \mu(z_{t-\nicefrac{1}{2}} - z_{t-1})\! + \! \alpha\beta(\hat{g}_{t+1} \!-\! \hat{g}_{t})\!\label{eq:optimism} \!
&&\text{\emph{(optimism)}} 
}
 If the gradient predictions $\{\hat{g}_{t+1}\}_{t\geq 0}$ are accurate $\hat{g}_{t+1} = g_{t+1}$, the optimistic update rule can accelerate. The policy updates are \textbf{extrapolations} based
on predictions of next surrogate objective in the sequence. Using $u_{t-1} = g_t$ we obtain the predictor-corrector approach \citep{predictor_corrector}. But in RL, agents generally do not have $g_t$, so the distance to the true gradient $\| g_t -u_{t-1}\|_*$ will depend on how good the prediction $\hat{g}_t$ was at the previous iteration $\|g_t - \hat{g}_t\|_*$, with $\|\cdot\|_*$ the dual norm.
 Since we have not computed $\pi_{t+1}$ at time $t$, and we do not have the prediction $\hat{g}_{t+1}$, existing methods perform the following techniques. 

 \paragraph{Lookahead}\emph{Model-based} policy optimization methods \citep{silver2017, Grill2020, Schrittwieser2019} use an update $u_{t}$ that anticipates the future performance, by looking one or multiple steps ahead ($h\geq 1$), using a model or simulator in place of the environment dynamics ($\hat{r}$, $\hat{P}$) to compute  $U^{(h)}_{t} \doteq \hat{\T}_{\pi_b}^h Q_t$. with $\pi_b$ a Tree Search policy, e.g. the greedy policy. With $Q_{t+1} \doteq \hat{\T}_{\pi_b} Q_t$, and $\hat{\E}_{\pi_b}$ the distribution under the model's dynamics
 \eqq{
  U^{(h)}_{t}  &= \hat{\T}_{\pi_b} Q_t + \hat{\T}_{\pi_b}^h Q_t- \hat{\T}_{\pi_b} Q_t 
  = Q_{t+1} + \gamma \hat{P}_{\pi_b}(\hat{\T}^{h-1} Q_t-  Q_t)  = Q_{t+1} +  \gamma\hat{\E}_{\pi_b}[U^{(h-1)}_{t} - Q_t]\nonumber
 }

 \paragraph{Extra-gradients} We interpret \emph{optimistic meta-learning} algorithms \citep{Flennerhag22, flennerhag2023optimistic} as extra-gradient methods, since they use the previous prediction $u_{t-1}$ as a proxy  
 to compute a \emph{half-step proposal} ${\pi}_{t+\nicefrac{1}{2}} =  \P_{\Pi} (\nabla \Omega)^{-1}(z_t + \alpha u_{t-1})$, which they use to obtain an estimate of the next gradient $\hat{g}_{t+\nicefrac{1}{2}} \doteq \nabla_{\pi} J({\pi}_{t+\nicefrac{1}{2}})$ (e.g.  for sample efficiency, using samples from a non-parametric model, like a reply buffer). Retrospectively, they adapt the optimistic update
 \eqq{
\! u_t  \!\doteq\! \beta \hat{g}_{t+\! 
\nicefrac{1}{2}}\! +\! \mu[{u}_{t-1} \!- \! \beta \hat{g}_{t}] \!\implies\! z_{t+\nicefrac{1}{2}} \!=\! z_t \!+\! \mu(z_{t-\! \nicefrac{1}{2}}\! -\! z_{t-1})\! + \! \alpha\beta(\hat{g}_{t+\! \nicefrac{1}{2}}- \!\hat{g}_{t})\! \!
&&\text{\emph{(extra-grad)}}\! \label{eq:extragrads}
}
A new target policy should also be recomputed ${\pi}_{t+1} =  \P_{\Pi} (\nabla \Omega)^{-1}(z_t + \alpha u_{t})$, but practical implementations \citep{Zahavy20, Flennerhag22} omit this, and resort to starting the next iteration from the half-step proposal ${\pi}_{t+1} \doteq {\pi}_{t+\nicefrac{1}{2}}$. 
Alg.~\ref{alg:acc_pi} summarizes the procedure.

\subsection{Towards a practical Accelerated Policy Gradient algorithm}\label{sec:optimistic_policy_gradients}
More commonly used in practice are neural, or log-linear parametrizations for actors, and equivalent parametrizations of gradient-critics (e.g., \citet{impala, Schulman15, Schulman17, Abdolmaleki18, Tomar20, Zahavy20, Flennerhag22, Hessel2021}).  
Consider a parameterized softmax policy class $\Pi_{\Theta} \!\doteq\! \big\{\pi_\theta | \pi_{\theta}(s,a) \!\doteq\! \nicefrac{\exp z_\theta(s,a)}{\sum_a \exp z_\theta(s,a)},\!\forall s\!\in\!\S, a\!\in\!\A, \theta\!\in\! \Theta \!\subset\! \mathbb{R}^m\big\}$, with $\pi_{\theta}\! \in \!\Pi_{\Theta}$, and $z_{\theta}$ a differentiable logit function.
Let $u_{\eta}$ represent a parametric class of policy updates, with parameters $\eta \in \R^{m^\prime}$, which we discuss momentarily.

\paragraph{Base algorithm} We recast the policy search in Eq.~\ref{eq:local surrogate objective} over policy parameters $\theta$
\eqq{
\theta_{t+1} &\doteq \arg\max_{\theta\in \Theta} \ell_{t}(\pi_{\theta}, u_{\eta_t})\qquad \ell_{t}(\pi_{\theta}, u_{\eta}) \doteq  \langle \pi_{\theta}, u_{\eta}\rangle -  \nicefrac{1}{\alpha} \KL_{[d_{\pi_{\theta_t}}]}(\pi_{\theta}, \pi_{\theta_t})  \label{eq:surrogate_objective} 
}
Using function composition, we write the policy improvement step using a parametrized recursive algorithm $\pi_{\theta_{t+1}} = \varphi(\eta_{t-1}, \pi_{\theta_t})$ with $\eta$ the algorithm's (meta-)parameters. We assume the $\varphi(\eta, \cdot)$ is differentiable and smooth w.r.t. $\eta$.
If Eq.~\ref{eq:surrogate_objective} can be solved in closed form, an alternative is to compute the non-parametric closed-form solution $\pi_{t+1} \propto {\pi_{\theta_t}\exp{\alpha u_{\eta_t}}}$ and (approximately) solve the projection $\theta_{t+1} \doteq \arg\min_{\theta\in \Theta} \KL(\pi_{\theta}, \pi_{t+1})$. For both approaches, we may use $h\geq 1$ gradient steps, to solve on Eq.~\eqref{eq:surrogate_objective}
\eqq{
\theta^{k+1}_t = \theta^{k}_{t} + \xi y_{t+1} \qquad y_{t+1} \doteq \nabla_{\theta} \ell_{t}(\pi_{\theta^k_t}, u_{\eta_t}) \qquad  \forall k \in [0..h), \theta^{0}_{t} \doteq \theta_{t}\qquad  \theta_{t+1}\doteq \theta^{h}_{t}
\label{eq:surrogate_objective_optimization} 
}
with $\xi$ a parameter step size. 
By function compositionality, we have $ \nabla_{\theta} \ell(\pi, g) = \nabla_{\theta} \pi_{\theta}^\top \nabla_{\pi} \ell(\pi, g)$.
This part of the gradient $\nabla_{\theta} \pi_{\theta}(s,a)=\pi_{\theta}(s,a)\nabla_{\theta} \log \pi_{\theta}(s,a)$ is generally not estimated,  available to RL agents, and computed by backpropagation.
Depending on how the other component $\nabla_{\pi} \ell(\pi, g)$ is defined, in particular $u_{\eta_t}$, we may obtain different algorithms. Generally, this quantity is expensive to estimate accurately in the number of samples for RL algorithms.

\begin{wrapfigure}{r}{0.55\textwidth} 
\begin{minipage}{0.55\textwidth} 
\vspace{-23pt}
\begin{algorithm}[H]
\caption{Accelerated Policy Gradients (in practice)
}
\label{alg:meta}
\begin{algorithmic}
{\footnotesize
 \STATE {\bfseries input: }$(\theta_0, \eta_0)$,  predictions $\{Q_{t}\}_{t\geq0}$
  \\
  \FOR{$\text{each iteration } t = 1,2 \dots$}
        \STATE Sample from $\pi_{\theta_{t}}$ \& store in buffer $\mathcal{B}_{\hat{d}, \pi_t}$
         \STATE \emph{\# policy improvement}
         \STATE Update ${\theta_{t+1}}$ with Eq.~\ref{eq:surrogate_objective practice}
         \STATE \emph{\# acceleration}
         \STATE  Compute $\pi_{{t+2}} \!\propto\! \pi_{\theta_{t+1}} \exp \alpha Q_{t+1}$  or $\pi_{\theta_{t+2}} \!= \!\arg\max_{\theta} \ell(\pi_{\theta}, Q_{t+1})$ with $\ell$ from Eq.~\ref{eq:surrogate_objective practice} using samples from $\mathcal{B}_{\hat{d}, \pi_{t+1}}$
         \STATE Update $\eta_{t+1}$ with Eq.~\ref{eq:meta_objective} and  $\mathcal{B}_{\hat{d}, \pi_{t+1}}$
    \ENDFOR
    }
\end{algorithmic}
\end{algorithm}
\end{minipage}
\vspace{-10pt}
\end{wrapfigure}

In order to admit an efficient implementation of the parametrized surrogate objective in Eq.~\ref{eq:surrogate_objective}, we only consider separable surrogate parametrizations over the state space. We resort to sampling experience under the empirical distribution $\hat{d}$ and the previous policy $\pi_{\theta_t}$, and we replace the expectation over states and actions in the gradient with an empirical average over rollouts or mini-batches $\mathcal{B}_{\hat{d}, \pi_t} \doteq \dot \{(S^{0}, A^{0}), (S^{1}, A^{1})\dots\}$. Leveraging this compositional structure, the algorithms we consider use weighted policy updates $u_{\eta}(s, a) = \hat{d}(s) U_{\eta}(s, a)$
\eqq{
\ell_{t}(\pi_{\theta}, U_{\eta}) \doteq  \E_{\mathcal{B}_{\hat{d}, \pi_t}}[\nicefrac{\pi_{\theta}(A|S)}{\pi_{\theta_t}(A|S)} U_{\eta}(S,A) -  \nicefrac{1}{\alpha} \KL(\pi_{\theta}(\cdot|S), \pi_{\theta_t}|(\cdot|S)]  \label{eq:surrogate_objective practice} 
}

\begin{example}{\textbf{(A non-optimistic algorithm)}} 
Under this definition, the standard actor-critic algorithm uses $U_{\eta} \doteq Q_{w}$ and updates $w$ with semi-gradient temporal-difference (TD) algorithms toward a target $Q_{t+1}$, typically bootstrapped from $Q_{w_t}$. Let  $\zeta$ be a step size. The TD objective denoted as $w_{t+1} \doteq \arg\min_{w\in\R^{m^\prime}}  f_t(\eta,{Q}_{{t+1}})$ is a surrogate objective for  value error \citep{Patterson2021AGP}
 \eqq{
  f_t(w,{Q}_{{t+1}})\! \doteq\! \E_{\mathcal{B}_{\hat{d}, \pi_t}}[ \nicefrac{1}{2}\big(\lceil{Q}_{{t+1}}\rfloor(S,A) - Q_{w}(S,A)\big)^2] \!+\! \nicefrac{1}{2\zeta}\|w \!-\! w_t\|_2^2 \label{eq:standard_critic}
 }
 \end{example}

\paragraph{Acceleration}
We replace the standard policy update with an optimistic decision-aware update, retrospectively updated using the objective
\eqq{
f_t(\eta,{Q}_{{t+1}})\! \doteq \!  \beta\ell_{t+1}(\pi_{{t+2}},  Q_{t+1} ) \! + \!  \mathcal{B} [\ell_t(\varphi(\eta, \pi_{\theta_t}), U_{\eta}) \! -\!  \beta\ell_{t}(\pi_{\theta_{t+1}},  Q_{t} )] \label{eq:meta_obj}
}
where we used the same notation as before, $\pi_{\theta_{t+1}} \doteq \varphi(\eta_{t-1}, \pi_{\theta_t})$, to denote that the policy improvement step uses a parametrized recursive algorithm with parameters $\eta_{t-1}$, and $\ell$ defined in Eq.~\ref{eq:surrogate_objective practice}. The optimal solution to $\ell_t(\varphi(\eta_{t-1}, \pi_{\theta_t}))$ is $\pi_{\theta_{t+1}} \!=\! \P_{\Pi_{\Theta}} \pi_{t+1}$, with $\pi_{t+1} \doteq \arg\max_{\pi} \ell_t(\pi, U_{\eta_{t-1}}) \!\propto\! \pi_{\theta_t} \exp( \alpha U_{\eta_{t-1}})$, and the optimal next-iteration target is $\pi_{t+2} \doteq \arg\max_{\pi} \ell_{t+1}(\pi, Q_{t+1}) \!\propto \!\pi_{\theta_{t+1}} \exp(\alpha Q_{t+1})$, which we may be also approximate using Eq.\ref{eq:surrogate_objective_optimization}.

Since $\ell_t(\pi_{\theta_{t+1}}, U_{\eta_{t-1}})  = \KL(\pi_{\theta_t}, \pi_{\theta_{t+1}})$, $\ell_{t+1}(\pi_{{t+2}},  \beta Q_{t+1} )- \ell_{t}(\pi_{\theta_{t+1}},  \beta Q_{t})=\KL(\pi_{\theta_{t+1}}, \pi_{{t+2}})$, the objective in Eq.~\ref{eq:meta_obj} is captured in the left-hand side of the generalized Pythagorean theorem 
\eqq{
& \!\KL(\pi_{\theta_t}, \pi_{\theta_{t+1}}) \!+\! \KL(\pi_{\theta_{t+1}}, \pi_{{t+2}}) = \KL(\pi_{\theta_t}, \pi_{t+2}) +\!\langle \nabla_{\pi}\KL(\pi, \pi_{t+2})|_{\pi=\pi_{\theta_{t+1}}}, 
\pi_{\theta_t} - \pi_{\theta_{t+1}}
\rangle \nonumber
 }
 By cosine law, if $\langle \nabla_{\pi}\KL(\pi, \pi_{t+2})|_{\pi=\pi_{\theta_{t+1}}}, 
\pi_{\theta_t} -\pi_{\theta_{t+1}})  
\rangle\geq 0$, then $\KL(\pi_{\theta_t}, \pi_{t+2})\geq \!\KL(\pi_{\theta_t}, \pi_{\theta_{t+1}}) \!+\! \KL(\pi_{\theta_{t+1}}, \pi_{{t+2}}) $, and $\pi_{\theta_{t+1}}$ was not the optimal projection of $\pi_{t+2}$, i.e. $\pi_{\theta_{t+1}} \neq \arg\min_{\pi\in\Pi} KL(\pi, \pi_{t+2})$. Therefore, we can move $\eta$ to relax $\KL(\pi_{t}, \pi_{t+2})$ by minimizing $\langle \nabla_{\pi}\KL(\pi, \pi_{t+2})|_{\pi=\pi_{\theta_{t+1}}}, 
\pi_{\theta_t} -\varphi(\eta, \pi_{\theta_t})  
\rangle$. 
 We use a first-order method, which linearizes this objective in the space of parameters $\eta$
\eqq{
 f_t(\eta, Q_{t+1})\!\doteq\! \langle \eta, \nabla_{\eta} \varphi(\eta_t, \pi_{\theta_t})^\top \nabla_{\pi} \KL(\pi_{\theta_{t+1}}, \pi_{t+2})\rangle \!+\! \nicefrac{1}{2\zeta}\|\eta \!-\! \eta_t\|_2^2 \label{eq:meta_objective}
}
where $\pi_{t+2}$ depends on $Q_{t+1}$.
Alg.~\ref{alg:meta} summarizes the procedure.

Next, we empirically study: (i) the effect of grounded meta-optimization targets ${Q}_{t+1}$ based on true optimistic predictions ${Q}_{t+1} \doteq Q_{\pi_{\theta_{t+1}}}$, and (ii) using self-supervised, inaccurate predictions---obtained with another estimator: ${Q}_{t+1} \doteq Q_{w_{t+1}}$, with $w$ learned separately with TD. 
We leave to future work the exploration of other ways of adding partial feedback to ground the bootstrap targets.

\subsubsection{Illustrative empirical analysis}\label{sec:empirical_analysis}

In this section, we investigate acceleration using optimism for online policy optimization, in an illustrative task. We mentioned one option for computing optimistic predictions $\{Q_{t}\}_{t\geq0}$ is using a model or simulator.
Consequently, in Sec.~\ref{subsec:Optimism with multi-step forward search}, we begin with a brief study on the effect of the lookahead horizon on the optimistic step, in order to understand the acceleration properties of multi-step Tree Search algorithms, and distinguish between two notions of optimism.
Thereafter, in Sec.~\ref{subsec:Optimistic policy gradients}, we consider the accelerated policy gradient algorithm we designed in Sec.~\ref{sec:optimistic_policy_gradients} (summarized in Alg.~\ref{alg:meta}), and investigate emerging properties for several choices of policy targets $\pi_{t+2}$ obtained with optimistic predictions $\{Q_{t+1}\}_{t\geq0}$.  

\begin{wrapfigure}{r}{0.2\textwidth} 
\vspace{-30pt}
    \centering
    \includegraphics[width=0.15\textwidth]{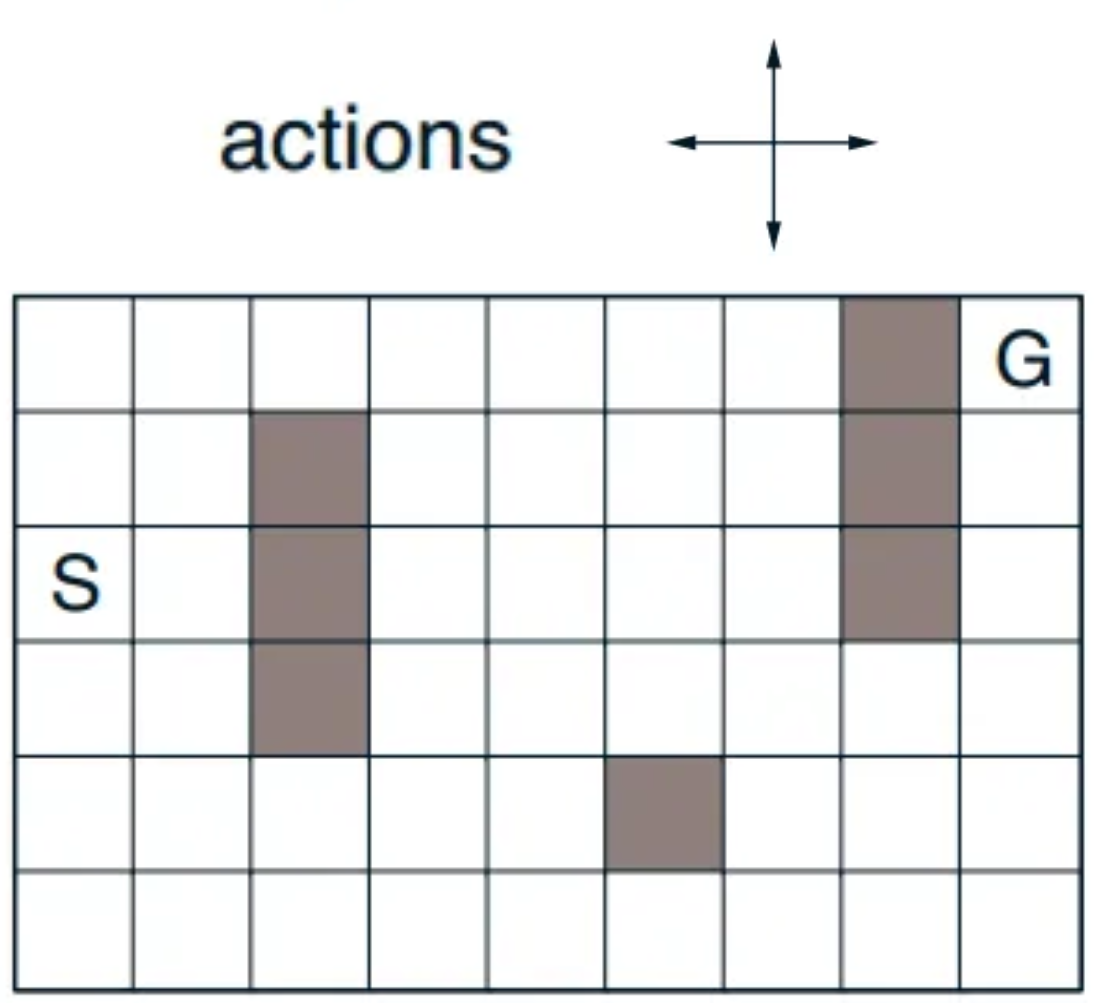}
    \vspace{-35pt}
\end{wrapfigure}

\textbf{Experimental setup} \quad In both  experiments, we use the discrete navigation task from \citep{rl_book}, illustrated aside (details in Appendix~\ref{apend:empirical_analysis}). 

\subsubsection{Optimism with multi-step forward search}\label{subsec:Optimism with multi-step forward search}
For this illustration, we use exact simulators. The only source of  inaccuracy stems from the depth truncation from using a fixed lookahead horizon $h$.
We use this experiment to show the difference between: (i) optimism within the local policy evaluation problem ($U_{t} \doteq \T^h_{\pi_t} {Q}_t$), and (ii) optimism within the global maximization problem ($U_{t} \doteq \T^h {Q}_t$).

 \textbf{Algorithms} \quad We consider an online AC algorithm, with forward planning up to horizon $h$ for computing the search values $U_{t}  \doteq \T_{\pi_b}^h {Q}_{w_t}$, bootstrapping at the leaves on ${Q}_{w_t}$, trained with using Eq.~\ref{eq:standard_critic}, and $\pi_b$, a tree-search policy. We optimize the policy $\pi_{\theta}$ online, using $h=1$ gradient steps on Eq~\ref{eq:surrogate_objective practice}: $\theta_{t+1}= \theta_t+ \beta \nabla_\theta \log \pi_{\theta_t} (A|S){\Adv}_{t}(S,A)$, with actions sampled online and $\Adv_{t} \doteq  {U}_{t} - V_{t}$ the $h$-step advantage, where $V_{t}(S) \doteq \E_{\pi_{\pi_b}(\cdot|S)}[{U}_{t}(S,A)]$.

\begin{figure}[t]
\centering
 \subfigure[\centering\footnotesize Optimistic evaluation]{\label{fig:fw_search:b}\includegraphics[width=0.3\textwidth]{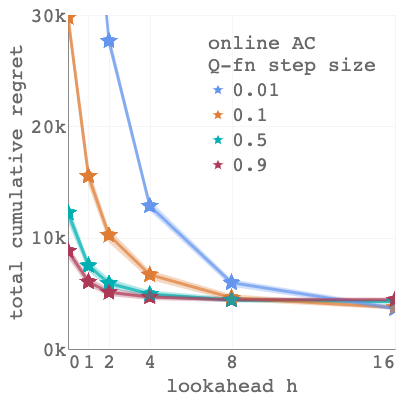}}
 \hfill
 \subfigure[\centering\footnotesize Optimistic improvement]{\label{fig:fw_search:c}\includegraphics[width=0.3\textwidth]{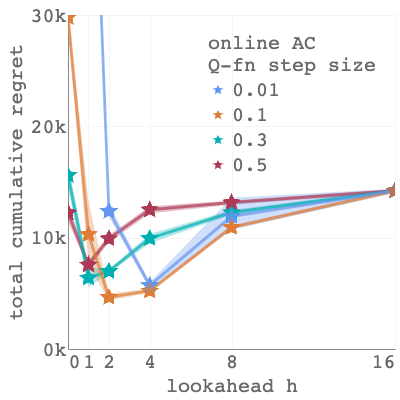}}
  \hfill
    \subfigure[\centering\footnotesize Difference between (a) and (b)]{\label{fig:fw_search:a}\includegraphics[width=0.3\textwidth]{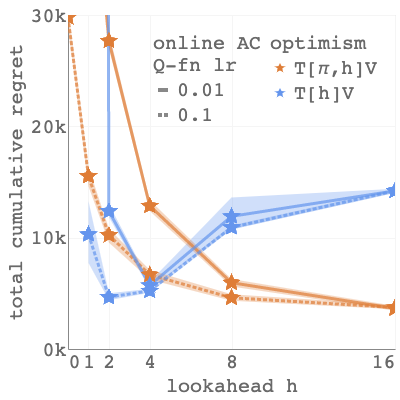}}
    \vspace{-5pt}
\caption{\footnotesize \textbf{Optimism with extra steps of forward search with a simulator.} x-axis: lookahead horizon $h$; y-axis: total cumulative regret $\sum_{k\leq t} J(\pi^*)- J(\pi_k)$. Lookahead targets $r(S,A)+\gamma U_{t}(S^\prime, A^\prime)$ are used for:
\textbf{(a) Optimistic evaluation}  $U_{t}\doteq \T^{h}_{\pi_t} {Q}_{w_t}$. Increasing the optimistic lookahead horizon $h$ helps, and a horizon $h = 0$ is worst. Colored curves denote the step size $\zeta$ used to learn the parameter vector $w$ of $Q_w$ with online TD($0$). The step size controls the quality of the gradient via the accuracy of the search values $U_{t}$ (more details in the main text). Shades denote confidence intervals over $10$ runs. \textbf{(b) Optimistic improvement} $U_{t}\doteq \T^h {Q}_{w_t}$. Intermediate values of the optimistic lookahead horizon $h$  trade off accumulating errors for shorter horizons.
 \textbf{(c) Comparison between the two notions of optimism}  {\color{retro_orange}  local}---evaluation within the current prediction problem, and {\color{cornflower_blue} global}---improvement within the optimization problem, 
 for two step sizes.}
     \label{fig:fw_search}
     \vspace{-10pt}
\end{figure}

\textbf{\emph{Relationship between acceleration and optimistic lookahead horizon}}\quad
 We use a multi-step operator in the optimistic step, which executes Tree-Search up to horizon $h$. For the tree policy $\pi_b$, we distinguish between: (i) extra policy evaluation steps with the previous policy, ${U}_{t}\doteq \T_{\pi_t}^{h} {Q}_{w_t}$ (Fig~\ref{fig:meta:a}), and (ii) extra greedification steps, ${U}_{t}\doteq \T^{h} {Q}_{w_t}$ (Fig~\ref{fig:meta:b}). The policy is trained online with $\theta_{t+1} = \theta_t + \xi y_t$, s.t. $\E_{\hat{d}, \pi_t}[y_t]\doteq\E_{\mathcal{B}_{\hat{d}, \pi_t}}[{\Adv}_{t}(S,A)\nabla_\theta \log\pi_{\theta_t}(A|S))]$, with ${\Adv}_{t}= {U}_{t} - V_{t}$, where $V_{t}(S) \doteq \E_{\pi_{t}(\cdot|S)}[{U}_{t}(S,A)]$. The advantage function ${\Adv}_{t}$ uses search values ${U}_{t}$, and critic parameters $w$ trained online with Eq.~\ref{eq:standard_critic} from targets based on the search values $r(S,A) + \gamma {U}_{t}(S^\prime,A^\prime)$.

\textbf{Results \& observations} \quad Fig.~\ref{fig:fw_search:a} shows the difference between {\color{cornflower_blue} optimistic improvement}---the gradient prediction has foresight of future policies on the optimization landscape, and {\color{retro_orange} optimistic evaluation}---the gradient prediction is a refinement of the previous gradient prediction toward the optimal solution to the local policy improvement sub-problem. 
As Fig.~\ref{fig:fw_search:b} depicts, more lookahead steps with {\color{retro_orange} optimistic evaluation}, can significantly improve inaccurate gradients, where accuracy is quantified by the choice of $\zeta$, the Q-fn step size for $w$.
Thus, for $\pi_b\doteq \pi_t$, increasing $h\to\infty$, takes the optimistic step with the exact (functional) policy gradient of the previous policy, ${U}_{t}= \T_{\pi_b}^h Q_{w_t}= \T_{\pi_t}^h Q_{w_t} \overset{h\to\infty}{\longrightarrow}Q_{\pi_t}$.
As Fig.~\ref{fig:fw_search:c} shows, the optimal horizon value for {\color{cornflower_blue} optimistic improvement} is another, one that trades off the computational advantage of extra depth of search, if this leads to accumulating errors, as a result of depth truncation, and bootstrapping on inaccurate values at the leaves, further magnified by greedification.

\subsubsection{Accelerated policy optimization with optimistic policy gradients}\label{subsec:Optimistic policy gradients}

We now empirically analyze some of the properties of the practical meta-gradient based adaptive optimistic policy gradient algorithm we designed in Sec.~\ref{sec:optimistic_policy_gradients} (Alg.~\ref{alg:meta}).

\textbf{\emph{(i) Acceleration with optimistic policy gradients}} \quad
We first remove any confounding factors arising from tracking inaccurate
target policies ${\pi}_{t+2}$ in Eq.\ref{eq:meta_objective}, and resort to using the true gradients of the post-update performance of $\pi_{\theta_{t+1}}$, $Q_{t+1}\! \doteq \!Q_{\pi_{\theta_{t+1}}}$,  
 but distinguish between two kinds of lookahead steps: {\color{sea_green}(a) \emph{parametric}}, or {\color{cornflower_blue}(b) \emph{geometric}}.
 This difference is indicative of the farsightedness of the optimistic prediction.
In particular, this distinction is based on the policy class of the target, whether it be a  {\color{sea_green}(a) parametric policy target} $ 
 \pi_{{\theta}_{t+2}}$, obtained using $h$ steps on Eq.~\ref{eq:surrogate_objective_optimization}, with $y_{t+1}\doteq \nabla_{\theta} \ell_t(\pi_{\theta^k_{t+1}}, Q_{t+1})\forall k\geq h$, or a {\color{cornflower_blue}(b) non-parametric policy target}, ${\pi}_{t+2}\propto \pi_{\theta_{t+1}} \exp \alpha Q_{{t+1}}$. The results shown are for $h=1$, and $\alpha= 1$.

\begin{figure}[t]
 \subfigure[\footnotesize Expert targets]{\label{fig:meta:a}\includegraphics[width=0.33\textwidth]{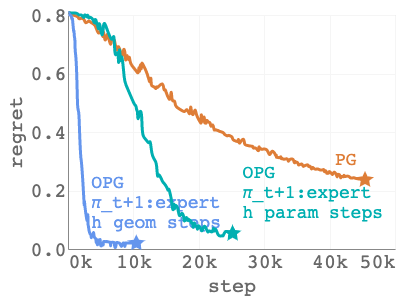}}
 \subfigure[\footnotesize Target predictions]{\label{fig:meta:b}\includegraphics[width=0.33\textwidth]{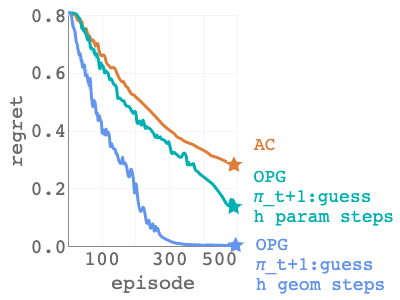}}
 \subfigure[\footnotesize Hyperparameter sensitivity]{\label{fig:meta:c}\includegraphics[width=0.33\textwidth]{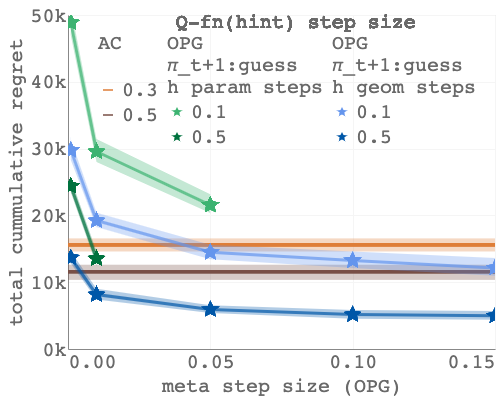}}
 \vspace{-5pt}
\caption{\footnotesize
 \textbf{Accelerated policy optimization with optimistic policy gradients} \textbf{(a)} x-axis: number of steps,
 y-axis: regret $J(\pi^*)- J(\pi_t)$. Different colored curves denote: {\color{retro_orange} standard PG}, \emph{optimistic policy gradients (OPG)} - with {\color{sea_green} parametric target policies}, and {\color{cornflower_blue} non-parametric target policies}, trained with meta-gradient learning from  optimistic predictions using the true post-update gradients. \textbf{(b)}
 x-axis: number of episodes,  y-axis: regret. \emph{Optimistic policy gradients (OPG)} are meta-learned from inaccurate optimistic predictions using Q-fn estimations. \textbf{(c)} Hyper-parameter sensitivity curves. x-axis: meta-learning rate for $\eta$, y-axis: total cumulative regret $ \sum_{k\leq t} J(\pi^*) - J(\pi_k)$. The plot shows \emph{optimistic policy gradients} meta-learned from  inaccurate optimistic predictions. Different tones depict different accuracies of the optimistic prediction, indirectly quantified via the optimistic Q-fn's step size. Straight lines show a baseline {\color{retro_orange} standard AC}. Shades denote confidence intervals over $10$ runs.}
     \label{fig:meta}
     \vspace{-5pt}
\end{figure}

\textbf{Results \& observations} \quad When the meta-optimization uses an adaptive optimizer (Adam \citep{adam}), Fig~\ref{fig:meta:a} shows there is acceleration when using targets ${\pi}_{t+2}$ one step ahead of the learner {\color{sea_green} parametric}, or {\color{cornflower_blue} geometric}. The large gap in performance between the two optimistic updates owes to the fact that target policies that are one {\color{cornflower_blue} geometric} step ahead correspond to steepest directions of ascent, and consequently, may be further ahead of the policy learner in the space of parameters, leading to acceleration. Additional results illustrating sensitivity curves to hyperaparameters are added in Appendix~\ref{apend:empirical_analysis}. When the meta-optimization uses SGD, the performance of the meta-learner algorithms is slower, lagging behind the PG baseline, but the ordering over the optimistic variants is maintained (Fig~\ref{apend:fig:meta_pg:c} in Appendix~\ref{apend:empirical_analysis}), which indicates that the correlation between acceleration and how far ahead the targets are on the optimization landscape is independent of the choice of meta-optimizer.

\textbf{\textbf{\emph{(ii) How target accuracy impacts acceleration}}}\quad
Next, we relax the setup from the previous experiment, and use inaccurate   predictions $Q_{t+1}\approx Q_{\pi_{\theta_{t+1}}}$, instead of the true post-update gradients. In particular, we resort to online sampling under the empirical on-policy distribution $\hat{d}$, and use a standard Q-fn estimats  to track the action-value of the most recent policy ${Q}_{w_{t+1}}\approx{Q}_{\pi_{\theta_{t+1}}}$ using Eq.~\ref{eq:standard_critic}, with TD(0): $w_{t+1}= w_t - \zeta [r(S,A)+ \gamma Q_{w_t}(S,A)- Q_{w_t}(S,A)] \nabla_w Q_{w_t}(S,A)$, with step size $\zeta$. 
With respect to the policy class of the targets, we experiment with the same two choices {\color{sea_green} (a) \emph{parametric} $\pi_{\theta_t+2}$}, or {\color{cornflower_blue} (b) \emph{non-parametric} $\pi_{t+2}$ }. Targets are ahead of the optimistic learner, in {\color{sea_green} (a) \emph{parameter}} steps, for the former, and {\color{cornflower_blue} geometric} steps for the latter.

\textbf{Results \& observations}\quad Even when the target predictions are inaccurate,
Fig.\ref{fig:meta:b} shows that optimistic policy ascent directions distilled from lookahead targets that use these predictions can still be useful (meta-optimization uses Adam, although promising results are in Appendix~\ref{apend:empirical_analysis} also for meta-optimization with SGD).  {\color{cornflower_blue} Non-parametric targets}, ahead in the optimization, show similar potential as when using true optimistic predictions.  Fig~\ref{fig:meta:c} illustrates the total cumulative regret (y-axis) stays consistent across different levels of accuracy of the optimistic predictions used by the targets, which is quantified via the Q-fn step sizes ($\zeta$), and indicated by different tones for each algorithm. 
As expected, we observe {\color{sea_green} parametric} targets to be less robust to step size choices, compared to {\color{cornflower_blue} non-parametric} ones, analogous the distinct effect of non-covariant gradients vs natural gradients.

\section{Concluding remarks}\label{sec:discussion}
We presented a simple template for accelerating policy optimization algorithms, and connected seemingly distinct choices of algorithms: model-based policy optimization algorithms, and optimistic meta-learning. We drew connections to well-known accelerated universal algorithms from convex optimization, and investigated some of the properties of acceleration in policy optimization. We used this interpretation to design an optimistic PG algorithm based on meta-gradient learning, highlighting its features empirically.

\textbf{Related work}\quad
We defer an extensive discussion on related work to the appendix. The closest in spirit to this acceleration paradigm we propose is the predictor-corrector framework, used also by \citet{predictor_corrector}. The most similar optimistic  algorithm for policy optimization is AAPI \citep{Hao2020}. Both analyze optimism from a smooth optimization perspective, whereas we focus the analysis on Bellman-operators and PI-like algorithms, optimistic update rules, thus allowing the unification.
We extend the empirical analysis of \citet{Flennerhag22}, who only focused on meta-learning hyperparameters of the policy gradient, and used optimistic update rule in parameter space, which is less principled and lacks guarantees.
Other meta-gradient algorithms \citep{meta_critic, meta_return, meta_loss, Xu2020} take to more empirical investigations.
We focused on understanding the core principles common across methods, valuable in designing new algorithms in this space, optimistic in spirit.

\textbf{Future work}\quad
We left many questions unanswered, theoretical properties, and conditions on guaranteed accelerated convergence. The scope of our experiments stops before function approximation, or bootstrapping the meta-optimization on itself. Conceptually, the idea of optimizing for future performance has applicability in lifelong learning, and adaptivity in non-stationary environments \citep{Flennerhag22, luketina2022metagradients, Chandak2020}.

\section*{Acknowledgements}
Veronica Chelu gratefully acknowledges support from FRQNT---Fonds de Recherche du Québec, Nature et Technologies, and IVADO.

\counterwithin*{theorem}{section}
\counterwithin*{definition}{section}
\counterwithin*{lemma}{section}
\counterwithin*{corollary}{section}
\counterwithin*{observation}{section}
\counterwithin*{remark}{section}

\setlength{\parskip}{0cm}
\setlength{\parindent}{0.5cm}
\setlength{\topsep}{0.35cm plus 0.2cm minus 0.2cm}

\makeatletter
\def\thm@space@setup{%
  \thm@preskip=0.45cm plus 0.3cm minus 0.16cm
  \thm@postskip=0.25cm plus 0.05cm minus 0.16cm
}
\makeatother
\clearpage
\appendix

\section*{Appendix}

\section{Convergence rates for policy gradient algorithms} 
\label{apend:related_work_context}
\begin{table}[tbh]
\centering
{\footnotesize
\def\arraystretch{1.3}
\begin{tabular}{c c c c c c c}
\textbf{Alg} & \textbf{Opt} &\textbf{$\mathbf{\Pi/\Pi_{\Theta}}$} &\textbf{$\mathbf{Q}$} & \textbf{Notes} & \textbf{$\mathbf{\mathcal{O}(T)}$} & \textbf{Reference} 
\\\hline\hline\hline
\multirow{1}{0.29in}{PI}&\multirow{1}{0.05in}{\ding{55}}&\multirow{1}{0.1in}{$\Pi$}&- & \multirow{1}{1.0in}{convex $\Pi$}&\multirow{1}{0.41in}{linear}   & \multirow{1}{1.7in}{\citet{Ye2011}}
\\ \hline
\multirow{1}{0.29in}{API}&\multirow{1}{0.05in}{\ding{55}} &\multirow{1}{0.1in}{$\Pi$} &\multirow{1}{0.3in}{$\mathbb{R}^{|\!\S\!|\!\times\!|\!\A\!|}$}& \multirow{1}{1.0in}{convex $\Pi$}   &\multirow{1}{0.41in}{linear} &\multirow{1}{1.7in}{\citet{scherrer2016improved}}
\\\hline
\multirow{1}{0.29in}{SoftPI} &\multirow{1}{0.05in}{\ding{55}} &\multirow{1}{0.1in}{$\Pi$} &- &\multirow{1}{1.0in}{convex $\Pi$} &\multirow{1}{0.41in}{linear}  &\multirow{1}{1.7in}{\citet{bhandari21a}}
\\
\hline
\hline \hline
\multirow{1}{0.29in}{GA}  & \multirow{1}{0.05in}{\ding{55}}&\multirow{1}{0.1in}{$\Pi_\Theta$} &- & \multirow{1}{1.0in}{log barrier reg.} &\multirow{1}{0.41in}{$\mathcal{O}(1\!/\!\sqrt{T})$ }&\multirow{1}{1.7in}{\citet{Agarwal19}}
\\ \hline
\multirow{1}{0.29in}{GA} & \multirow{1}{0.05in}{\ding{55}} &\multirow{1}{0.1in}{$\Pi_\Theta$}  &-&-&  \multirow{1}{0.41in}{$\mathcal{O}(1\!/\!{T})$ }&\multirow{1}{1.7in}{\citet{Mei20a}}
 \\\hline
\multirow{1}{0.29in}{GA} & \multirow{1}{0.05in}{\ding{55}} &\multirow{1}{0.1in}{$\Pi_\Theta$} &-& \multirow{1}{1.0in}{entropy reg} &  \multirow{1}{0.41in}{linear}&\multirow{1}{1.7in}{\citet{Mei20a}}
\\\hline
\multirow{1}{0.29in}{PGA} &\multirow{1}{0.05in}{\ding{55}} &\multirow{1}{0.1in}{$\Pi_{\Theta}$}&- &- &\multirow{1}{0.41in}{$\mathcal{O}(1\!/\!\sqrt{T})$ } &\multirow{1}{1.7in}{\citet{BhandariandRusso_globalguarantees}}
\\\hline
\multirow{1}{0.29in}{PGA}
 & \multirow{1}{0.05in}{\ding{51}}  &\multirow{1}{0.1in}{$\Pi_{\Theta}$}&\multirow{1}{0.3in}{$\mathbb{R}^{|\!\S\!|\!\times\!|\!\A\!|}$} &\multirow{1}{1.0in}{adaptive step size}    &\multirow{1}{0.41in}{$\mathcal{O}(1\!/\!\sqrt{T})$} &
\multirow{1}{1.7in}{\citet{predictor_corrector}}
\\\hline
\multirow{1}{0.29in}{PGA} &\multirow{1}{0.05in}{\ding{55}}&\multirow{1}{0.1in}{$\Pi_{\Theta}$}&\multirow{1}{0.3in}{$\mathbb{R}^{|\!\S\!|\!\times\!|\!\A\!|}$} &\multirow{1}{1.0in}{(non/)convex $\Pi$} &\multirow{1}{0.41in}{$\mathcal{O}(1\!/\!\sqrt{T})$ } &\multirow{1}{1.7in}{\citet{Agarwal19}}
\\
\hline 
 \multirow{1}{0.29in}{PGA} & \multirow{1}{0.05in}{\ding{55}}  &\multirow{1}{0.1in}{$\Pi_\Theta$} &\multirow{1}{0.3in}{$\mathbb{R}^{|\!\S\!|\!\times\!|\!\A\!|}$} &   &\multirow{1}{0.41in}{$\mathcal{O}(1\!/\!\sqrt{T})$} &
\multirow{1}{1.7in}{\citet{Shani2019}}
\\\hline
\multirow{1}{0.29in}{PGA}
 & \multirow{1}{0.05in}{\ding{55}}  &\multirow{1}{0.1in}{$\Pi_\Theta$}&\multirow{1}{0.3in}{$\mathbb{R}^{|\!\S\!|\!\times\!|\!\A\!|}$} &\multirow{1}{1.0in}{entropy reg.}    &\multirow{1}{0.41in}{$\mathcal{O}(1\!/\!{T})$} &
\multirow{1}{1.7in}{\citet{Shani2019}}
\\\hline
\multirow{1}{0.29in}{PGA}
 & \multirow{1}{0.05in}{\ding{55}}  &\multirow{1}{0.1in}{$\Pi_\Theta$}&\multirow{1}{0.3in}{$\mathbb{R}^{|\!\S\!|\!\times\!|\!\A\!|}$} &\multirow{1}{1.0in}{adaptive step size}    &\multirow{1}{0.41in}{linear} &
\multirow{1}{1.7in}{\citet{Khodadadian2021}}
\\ \hline
\multirow{1}{0.29in}{PGA}
 & \multirow{1}{0.05in}{\ding{55}}  &\multirow{1}{0.1in}{$\Pi_\Theta$}&- &\multirow{1}{1.0in}{adaptive step size}    &\multirow{1}{0.41in}{linear} &
\multirow{1}{1.7in}{\citet{BhandariandRusso_globalguarantees}}
\\ \hline
\multirow{1}{0.29in}{PGA}
 & \multirow{1}{0.05in}{\ding{55}}  &\multirow{1}{0.1in}{$\Pi_\Theta$}&\multirow{1}{0.3in}{$\mathbb{R}^{|\!\S\!|\!\times\!|\!\A\!|}$} &\multirow{1}{1.0in}{adaptive step size}    &\multirow{1}{0.41in}{linear} &
\multirow{1}{1.7in}{\citet{xiao2022convergence}}
\\ \hline
\multirow{1}{0.29in}{PGA}
 & \multirow{1}{0.05in}{\ding{55}}  &\multirow{1}{0.1in}{$\Pi_\Theta$}&\multirow{1}{0.3in}{$\mathbb{R}^{|\!\S\!|\!\times\!|\!\A\!|}$} &\multirow{1}{1.0in}{entropy reg.}    &\multirow{1}{0.41in}{linear} &
\multirow{1}{1.7in}{\citet{cen}}
\\ \hline
\multirow{1}{0.29in}{PGA}
 & \multirow{1}{0.05in}{\ding{55}}  &\multirow{1}{0.1in}{$\Pi_\Theta$}&- &\multirow{1}{1.0in}{entropy reg.}    &\multirow{1}{0.41in}{linear} &
\multirow{1}{1.7in}{\citet{bhandari21a}}
\\ \hline
\multirow{1}{0.29in}{PGA}
 & \multirow{1}{0.05in}{\ding{55}}  &\multirow{1}{0.1in}{$\Pi_\Theta$}&\multirow{1}{0.3in}{$\mathbb{R}^{|\!\S\!|\!\times\!|\!\A\!|}$} &\multirow{1}{1.0in}{strong reg.}    &\multirow{1}{0.41in}{linear} &
\multirow{1}{1.7in}{\citet{lan2022policy}}
\\\hline\hline
\multirow{1}{0.29in}{PGA}
 & \multirow{1}{0.05in}{\ding{55}}  &\multirow{1}{0.1in}{$\Pi_{\Phi^\top\Theta}$} &\multirow{1}{0.3in}{$\Phi^\top\mathbf{\eta}$}& &\multirow{1}{0.41in}{$\mathcal{O}(1\!/\!{T})$} &
\multirow{1}{1.7in}{\citet{Agarwal19}}
\\\hline
\multirow{1}{0.29in}{PGA}
 & \multirow{1}{0.05in}{\ding{55}}  &\multirow{1}{0.1in}{$\Pi_{\Phi^\top\Theta}$}&\multirow{1}{0.3in}{$\Phi^\top\mathbf{\eta}$} &\multirow{1}{1.0in}{adaptive step size}    &\multirow{1}{0.41in}{$\mathcal{O}(1\!/\!{T^{2\!/\!3}})$}  &
\multirow{1}{1.7in}{\citet{politex}}
\\ \hline
\multirow{1}{0.29in}{PGA} &\multirow{1}{0.05in}{\ding{51}}
&\multirow{1}{0.1in}{$\Pi_{\Phi^\top\Theta}$} &\multirow{1}{0.3in}{$\Phi^\top\mathbf{\eta}$} &\multirow{1}{1.0in}{adaptive step size}   &\multirow{1}{0.41in}{$\mathcal{O}(1\!/\!{T^{3\!/\!4}})$}   & \multirow{1}{1.7in}{\citet{Hao2020} }
\\ \hline
 \multirow{1}{0.29in}{PGA}&\multirow{1}{0.05in}{\ding{51}}
&\multirow{1}{0.1in}{$\Pi_{\Phi^\top\Theta}$} &\multirow{1}{0.3in}{$\Phi^\top\mathbf{\eta}$}
&\multirow{1}{1.0in}{adaptive step size}  
& \multirow{1}{0.41in}{$\mathcal{O}(1\!/\!{\sqrt{T}})$} & \multirow{1}{1.7in}{\citet{lazic21a}}
\\ \hline
\multirow{1}{0.29in}{PGA}
 & \multirow{1}{0.05in}{\ding{55}}  &\multirow{1}{0.1in}{$\Pi_{\Phi^\top\Theta}$}&\multirow{1}{0.3in}{$\Phi^\top\mathbf{\eta}$} &\multirow{1}{1.0in}{geom incr. step size}    &\multirow{1}{0.41in}{linear} &
\multirow{1}{1.7in}{\citet{chen22i}}
\\\hline
\multirow{1}{0.29in}{PGA}
 & \multirow{1}{0.05in}{\ding{55}}  &\multirow{1}{0.1in}{$\Pi_{\Phi^\top\Theta}$}&\multirow{1}{0.3in}{$\Phi^\top\mathbf{\eta}$} &\multirow{1}{1.0in}{geom incr. step size}    &\multirow{1}{0.41in}{linear} &
\multirow{1}{1.7in}{\citet{alfano2023linear}}
\\\hline
\multirow{1}{0.29in}{PGA}
 & \multirow{1}{0.05in}{\ding{55}}  &\multirow{1}{0.1in}{$\Pi_{\Phi^\top\Theta}$}&\multirow{1}{0.3in}{$\Phi^\top\mathbf{\eta}$} &\multirow{1}{1.0in}{geom incr. step size}    &\multirow{1}{0.41in}{linear} &
\multirow{1}{1.7in}{\citet{yuan2023linear}}
\\ \hline
\hline\hline
\end{tabular}
\vspace{2pt}
\caption{\textbf{Summary of previous work on rates of convergence for policy gradient algorithms} \textbf{(Columns)} \textbf{Alg}---algorithm used, \textbf{Opt}---whether it uses optimism, \textbf{$\mathbf{\Pi/\Pi_{\Theta}}$}---policy class, \textbf{$\mathbf{Q}$}---the space of the gradient-critic prediction used (if not using the true gradient-critic/Q-fn of the policy performance objective, \textbf{$\mathbf{Q_{\pi}}$}, in which case it is marked with $-$), \textbf{$\mathbf{\mathcal{O}(T)}$}---iteration complexity (finite sample analysis of convergence), as a function of number of iterations $\mathbf{T}$, \textbf{Notes}---other assumptions, limitations, observations. \textbf{(Algorithm)} the following abbreviations are used: PI---policy iteration, API---approximate policy iteration, SoftPI---soft policy iteration, GA---gradient ascent, PGA---projected gradient ascent (including (inexact) natural gradient ascent, (inexact) mirror ascent, (inexact) dual-averaging, (inexact) primal-dual views). \textbf{(Policy class)} the following abbreviations are used: $\Pi$---tabular with direct/natural policy parametrization, $\Pi_\Theta$---tabular with softmax policy parametrization, $\Pi_{\Phi^\top \Theta}$---log-linear policy parametrization, i.e. the softmax transform is applied on a linear parametrization, $\Phi^\top \Theta$, with $\Phi$---the feature representation, and with corresponding linear gradient approximation over the policy's features $Q = \Phi^\top \eta$, with $\eta$---parameter vector $\eta$.}
\label{table:all}
}
\end{table}
\clearpage

\section{Related work} 
\subsection{Optimism in policy optimization}\label{apend:Operator-view}

\paragraph{Problem formulation}{
The RL problem consists in finding a policy $\pi$ maximizing the discounted return---the policy performance objective:  $ J(\pi) \equiv \E_{S\sim\rho}[V_\pi(S)] = (1-\gamma)\mathbb{E}_{\pi, \rho}\big[\sum_{t\geq 0}\gamma^t R_t\big]$, where $V_\pi \in \R^{|\S|}$ is the value function, and $Q_\pi \in \R^{|\S|\times|\A|}$ the action-value function of a policy $\pi \in \Pi = \{ \pi \in \mathbb{R}^{|\S|\times |\A|}_+ | \sum_{a\in\A} \pi(s,a) = 1, \forall s \in\S\}$, s.t. $Q_\pi(s,a) \equiv \E_\pi \left[\sum_{t=0}^\infty \gamma^t R_t|S_0 = s, A_0=a\right]$, and $V_\pi(s) \equiv \E_\pi\left[Q(s,A)\right]$.
}

\subsubsection{Policy iteration}{
\paragraph{Policy iteration}{
The classic \textbf{policy iteration} algorithm repeats consecutive stages of (i) one-step greedy policy improvement w.r.t. a value function estimate 
\eqq{
\pi_{t+1} \in \G(V_{\pi_t}) = \{\pi: \T_\pi V_{\pi_t} = \T V_{\pi_t}\} \iff \pi_{t+1} = \arg\max_{\pi \in \Pi} \langle \nabla J(\pi_t),\pi\rangle = \langle {Q}_{t}, \pi\rangle_{d_{\pi_t}}
}
with ${\G}$ the greedy set of $V_{\pi_t}$,
followed by (ii) evaluation of the value function w.r.t. the greedy policy 
\eqq{
V_{\pi_{t+1}} = \lim_{h \to \infty} \T^{h}_{\pi_{t+1}} V_{\pi_t} \text{ or } Q_{\pi_{t+1}} = \lim_{h \to \infty} \T^{h}_{\pi_{t+1}} Q_{\pi_t}
}
}
\paragraph{Approximate policy iteration}{
Approximations of either steps lead to approximate PI (API) \citep{scherrer15a}, in which we replace the two steps above with 
\eqq{
\pi_{t+1} \in \G(V_{\pi_t}) = \{\pi: \T_\pi V_{\pi_t} \geq \T V_{\pi_t} - \eps_{t+1}\}
}
with $\eps_{t+1}$ a greedification and/or value approximation error. 
}
\paragraph{Soft policy iteration}{
Relaxing the greedification leads to \textbf{soft policy iteration}, or conservative policy iteration \citep{cpi}, called Frank-Wolfe by \citet{bhandari21a}. The minimization problem decouples across states to optimize a linear objective over the probability simplex
\eqq{
\pi_{t+1} = (1-\alpha) \pi_t + \alpha \pi^+_{t+1} \text{ with
 } \pi_{t+1}^+ 
= \arg\max_{\pi\in\Pi} \langle {Q}_{\pi_t}, \pi\rangle_{d_{\pi_t}} }
for $\alpha\in [0,1]$, a (possibly time-dependent) step size, and $\langle \cdot, \cdot\rangle_{d}$ a state weighting that places weight $d(s)$ on any state-action pair $(s,a)$. 
}

\paragraph{Optimistic policy iteration (OPI)} \citep{bertsekas:ndp-book} relaxes the evaluation step instead to 
\eqq{
Q_{t+1} = (1-\lambda) Q_t + \lambda Q^+_{t+1} \text{, with } Q_{t+1}^+ = \T^h_{\pi_{t+1}} Q_t , \forall h \geq 0
}
with $\lambda\in [0,1]$.
}

\subsubsection{Policy gradients}
\paragraph{Projected Gradient Descent}{ Starting with some policy
$\pi\in \Pi$, an iteration of projected gradient ascent with step size $\alpha$ updates to the solution of the regularized problem
\eqq{
\pi_{t+1}&=\arg\max_{\pi} \langle \nabla J(\pi_t), \pi \rangle -\frac{1}{\alpha} \sum_{s\in \S} d_{\pi_t}(s) \sum_{a\in \A} (\pi(a|s) - \pi_t(a|s))^2
\\
&=\arg\max_{\pi} \langle Q_{\pi_t}, \pi \rangle_{d_\pi} -\frac{1}{\alpha} \|\pi - \pi_t\|^2_{2,d_{\pi_t}}
}
which is a first-order Taylor expansion of $J$ w.r.t. the policy's functional representation $\pi$ (see \citet{bhandari21a, BhandariandRusso_globalguarantees})
\eqq{
J(\pi^\prime) &= J(\pi) + \langle \nabla J(\pi), \pi^\prime - \pi\rangle - \mathcal{O}(\|\pi^\prime-\pi\|^2)
\\
&= J(\pi) + \langle Q_\pi, \pi^\prime - \pi\rangle_{d_{\pi}} - \mathcal{O}(\|\pi^\prime-\pi\|^2)
}
With per state decoupling, for specific values of $\alpha$ this yields a per state projection on the decoupled probability simplex
\eqq{
\pi_{t+1} &= \P^{\Pi}_{[d_{\pi_t}]} \pi_{t+2} = \arg\max_{\pi \in \Pi} \| \pi - \pi_{t+2}\|^2_{2, d_{\pi_t}} \text{ with }
\pi_{t+2} = \pi_t + \alpha Q_{\pi_t} 
}
with $ \|\cdot\|^2_{2, d_{\pi_t}}$ the weighted $L_2$-norm.
}

\paragraph{Mirror descent (MD)}{ 
Mirror descent adapts to the geometry of the probability simplex by using a non-Euclidean regularizer. The specific regularizer used in RL is the entropy function $H(\pi) \equiv \pi \log \pi$, such that the resulting mirror map is the $\log$ function. The regularizer decouples across the state space and captures the curvature induced by the constraint of policies lying on the policy simplex via the softmax policy transform.

Starting with some policy $\pi_t  \in \Pi_\Theta$, an iteration of mirror descent with step size $\alpha$ updates to the solution of a regularized problem
\eqq{
\pi_{{t+1}} &= \arg\max_{\pi\in\Pi} \langle \nabla J(\pi_t), \pi \rangle +\frac{1}{\alpha} \sum_{s\in \S} d_{\pi_t}(s) \KL(\pi(s),\pi_t(s)) 
\\
&=\arg\max_{\pi\in\Pi}  \langle  Q_{\pi_t}, \pi \rangle_{d_{\pi_t}} +\frac{1}{\alpha}\KL_{[d_{\pi_t}]}(\pi,\pi_t) 
}
which is known to be the exponentiated gradient ascent update $\pi_{t+1} = \frac{\pi_t \exp \alpha Q_{\pi_t}}{\sum_a \pi(a|\cdot) \exp \alpha Q_{\pi_t}(\cdot, a)}$ (obtained using the Lagrange approach, see \citet{Bubeck15}).

Using state decoupling, for specific values of $\alpha$ we may also write MD as a projection using the corresponding Bregman divergence for the mirror map $\nabla_\pi H(\pi)$ (cf. \citet{Bubeck15})
\eqq{
\pi_{t+1} &= \P^{\Pi, H}_{[d_{\pi_t}]} \pi_{t+2} = \arg\max_{\pi\in\Pi}\KL_{[d_\pi]}(\pi, \pi_{t+2}) \text{ with }
\\
\log\pi_{t+2} &= \log \pi_t +  \alpha Q_{\pi_t} - \log \sum_a \pi(a|\cdot) \exp \alpha Q_{\pi_t}(\cdot, a) 
}
}
\paragraph{Policy parametrization}{
For parametric policy classes the search written over policies,  translates into similar versions of the linear objective, except over policy parameters. 
Since the class of softmax policies can approximate stochastic policies to arbitrary precision, this is nearly (we can only come infinitesimally close to an optimal policy) the same as optimizing over the class $\Pi$.
}

\paragraph{Natural policy gradients (NPG)}{
The natural policy gradient (NPG) of \citet{kakade2001} applied to the softmax parameterization  is actually an instance of mirror descent for the entropy-based regularizer $H$.

Natural policy gradient is usually described as steepest descent in a variable metric defined by the Fisher information
matrix induced by the current policy \citep{kakade2001, Agarwal19}
\eqq{
\theta_{t+1} &= \theta_t + \alpha \F_\rho(\theta_t)^\dagger \nabla_{\theta_t} J(\pi_{\theta_t})
\\
\F_\rho(\theta_t) &= \E_{S\sim d_{\pi_{\theta_t}}, A\sim\pi_{\theta_t}}\left[\nabla_{\theta_t} \log \pi_{\theta_t} \nabla_{\theta_t} \log \pi_{\theta_t}^\top\right]
}
and is equivalent to mirror descent under some conditions \citep{raskutti2014information}.
}
  Cf. \citet{bhandari21a, analyUpdate2021}, the aforementioned base MD and NPG updates are closely related to the practical instantiations in TRPO \citep{Schulman15}, PPO \citep{Schulman17}, MPO \citep{Abdolmaleki18}, MDPO \citep{Tomar20}. All these algorithic instantiations use approximations for the gradient direction.

\subsubsection{Actor-critic methods} Generally, in RL, an agent only has access to partial evaluations of the gradient $\nabla_\pi J(\pi)$, and commonly these involve some sort of internal representation of the action-value function $Q_t \approx Q_{\pi_t}$. 

\paragraph{Natural actor-critic. MD with an estimated critic.}{ 
Consider a parameterized softmax policy class $\pi_{\theta} \in \Pi_{\Theta}$, with parameter vector $\theta$, and $Q_{\eta} \in \mathcal{F}_{\eta}$, with parameter vector $\eta$, s.t.  For the softmax policy class, this will be  $\log \pi_{\theta}$, for $\Pi_{\Theta} = \big\{ \pi_\theta  \big|  \pi_\theta(s,a)=\frac{\exp f_\theta(s,a)}{ \sum_{a^\prime\in\A}\exp f_\theta(s,a^\prime)}\forall s \in\S, a\in\A, \theta\in \mathbb{R}^m\big\}$, with $f_\theta$ a differentiable function, either tabular $f_\theta(s,a) = \theta_{s,a}$, log-linear $f_\theta(s,a) = \phi(s,a)^\top \theta$, with $\phi$ a feature representation, or neural ($f_\theta$-a neural network) parametrizations \citep{Agarwal19}. 

Written as a proximal policy improvement operator, at iteration $t$, starting with some policy $\pi_t \equiv \pi_{\theta_t}$. the next policy is the solution to the regularized optimization problem
\eqq{
\pi_{\theta_{t+1}} &= \arg\max_{\pi_\theta \in \Pi_\Theta} \langle Q_{w_{t}}, \pi_\theta\rangle_{d_{\pi_t}} - \frac{1}{\alpha}\KL_{[d_{\pi_t}]}(\pi_\theta, \pi_{t})
}
with $\alpha$ a (possibly time-dependent) step size, and $\langle \cdot, \cdot\rangle_{d}$, $\KL_{[d]}(\cdot, \cdot)$ indicates an additional state-weighting per state-action pair.

Using the connection between the NPG update rule with the notion of
compatible function approximation \citep{Sutton2000}, as formalized in \citep{kakade2001}, we may try to approximate the functional gradient using $w$
\eqq{
\F_\rho(\theta)^\dagger \nabla_\theta J(\pi_\theta) &= \frac{w}{1-\gamma} 
}
where $w$ are parameters of an advantage function $A_{w}$---which is the solution to the projection of $A_{\pi_\theta}$ on the dual gradient space of $\pi$, the space spanned by the particular feature representation that uses $\phi_t \equiv \nabla_{\theta} \log \pi_{\theta_t}$ as (centered) features
\eqq{
w_{t} &= \arg\min_w \E_{S\sim d_{\pi_{\theta_t}}, A\sim\pi_{\theta_t}} [(w^\top \phi_t(S,A) - \Adv_{\pi_{\theta_t}}(S,A))^2]
}
Similarly there is an equivalent version for Q-NPG considering possibly (un-centered) features ($\phi_{s,a}$, for $f_{\theta}(s,a) = \phi_{s,a}^\top \theta$) and projecting 
\eqq{
w_{t} &= \arg\min_w \E_{S\sim d_{\pi_{\theta_t}}, A\sim\pi_{\theta_t}} [(w^\top \phi_t(S,A) - Q_{\pi_{\theta_t}}(S,A))^2]
}
For both of them we can now replace the NPG parameter update with
\eqq{
\theta_{t+1} = \theta_t + \alpha w_{t}
}
}

\subsubsection{Forward search}
\paragraph{Multi-step policy iteration}{
The single-step based policy improvement used in the aforementioned algorithms, e.g., policy iteration, approximate PI, actor-critic methods, and its practical algorithmic implementations, is not necessarily the optimal choice. It has been empirically demonstrated in RL algorithms based on Monte-Carlo Tree Search (MCTS)\citep{BrownePWLCRTPSC12} (e.g., \citet{Schrittwieser2019, schmidhuber1987}) or Model Predictive Control (MPC), that multiple-step
greedy policies can perform conspicuously better.
Generalizations of the single-step greedy policy improvement include (i) $h$-step greedy policies, and (ii) $\kappa$--greedy policies. The former output the first optimal action out of a sequence of actions, solving a non-stationary $h$-horizon control problem:
\eqq{
\pi(s) \in \arg\max_{\pi_0} \max_{\pi_1, \dots \pi_{h-1}} \E^{\pi_0, \dots \pi_{h-1}} \left[\sum_{t=0}^{h-1} \gamma^t r(S_t, \pi_t(S_t))+ \gamma^h V(S_h)|S_0 = s \right]
}
equivalently described in operator notation as $\pi \in \G(\T^{h-1} V) \equiv \{\pi| \T_{\pi} T^{h-1} V\geq \T^h V\}$. 
A $\kappa$-greedy policy interpolates over all geometrically $\kappa$-weighted $h$-greedy policies $\pi \in \G(\T^\kappa V) \equiv \{\pi| \T^\kappa_{\pi} V\geq \T^\kappa V, \T^\kappa_{\pi} \equiv (1-
\kappa) \sum_{h=0}^\infty \kappa^h\T_\pi^{h+1}\}$. 
}
\paragraph{Multi-step soft policy iteration}{\citet{Efroni2018b} shows that when using soft updates with $h>1$
\eqq{
\pi_{t+1} = (1-\alpha) \pi_t + \alpha \pi^+_{t+1}, \pi^+_{t+1} \in \G(\T^{h-1} V) \equiv \{\pi| \T_{\pi} T^{h-1} V\geq \T^h V\}
}
policy improvement is guaranteed only for $\alpha = 1$, and when using
\eqq{
\pi_{t+1} = (1-\alpha) \pi_t + \alpha \pi^+_{t+1}, \pi^+_{t+1} \in \G(\T^{\kappa} V) \equiv \{\pi| \T^\kappa_{\pi} V\geq \T^\kappa V, \T^\kappa_{\pi} \equiv (1-
\kappa) \sum_{h=0}^\infty \kappa^h\T_\pi^{h+1}\}
}
policy improvement is guaranteed only for $\alpha \in [\kappa,1]$. This result appears in \citet{Efroni2018b}, and a more general version in \citet{Konda1999}.
}

\paragraph{Tree search}{Notable examples of practical algorithms with empirical success that perform multi-step greedy policy improvement are  AlphaGo and Alpha-Go-Zero \citep{silver2016,silver2017,SilverHMGSDSAPL16}, MuZero \citep{Schrittwieser2019}. There,
an approximate online version of multiple-step greedy improvement is implemented via Monte Carlo Tree Search
(MCTS) \citep{BrownePWLCRTPSC12}. In particular, \citet{Grill2020} shows that the tree search procedure implemented by AlphaZero is an approximation of the regularized optimization problem
\eqq{
\pi_{\theta_{t+1}} &= \arg\max_{\pi_\theta \in \Pi_\Theta} \langle  U_{t}, \pi\rangle_{d_{\pi_t}} - \frac{1}{\alpha_t}\KL_{[d_{\pi_t}]}(\pi_\theta, \pi_{t})
}
with $U_{t}$---the Tree Search values, i.e., those estimated by the search algorithm that approximates $\T^h Q_{w_{t}}$---$w$ parameters of the critic, with stochastic sampling of trajectories in the tree up to a horizon $h$, and bootstrapping on a Q-fn estimator at the leaves. For a full description of the algorithm, refer to \citet{silver2017}. The step size $\alpha_t$ captures the exploration strategy, and decreases the regularization based on the number of simulations.
}

\subsubsection{Meta-learning}
\paragraph{Optimistic meta-gradients}\quad
Meta-gradient algorithms further relax the optimistic policy improvement step to a parametric update rule $\pi_{\theta_{t+1}} \equiv \varphi_{\pi_{\theta_{t}}}({\eta_t})$,  e.g., $\theta_{t+1} = \theta_{t} + u_{\eta_t}$, when limited to a functional class of parametric GA update rules $u_{\eta} \in \mathcal{F}_{\eta}$. These algorithms implement adaptivity in a practical way, they project policy targets $\pi_{t+2}$ ahead of $\pi_{\theta_{t+1}}$
\eqq{
u_{\eta_{t+1}} &= \arg\min_{u_{\eta} \in \mathcal{F}_{\eta}} \KL_{[d_{\pi_{{t+1}}}]}(\pi_{\theta_{t+1}}, \pi_{t+2})   &&\text{\emph{(hindsight adaptation \& projection)}}
\label{eq:adaptation_step_pg2}
}
The targets can be parametric $\pi_{t+2} \equiv \pi_{{\theta}_{t+2}}$, initialized from ${\theta}^{(0)}_{t+1} = \theta_{t+1}$, and evolving for $h$ step further ahead of $\theta_{t+1}$, s.t. ${\theta}_{t+1}^{(k+1)} =  {\theta}^{k}_{t+1} + g^{k}_{t}, \forall k\leq h$, with $g^{k}_{t}$ representing predictions used by the bootstrapped targets. Alternatively, targets may be non-parameteric, e.g., ${\pi}_{t+2} \propto \pi_{\theta_{t+1}} \exp ({Q}_{t+1} - {Q}_{t})$, e.g., if
${Q}_{t+1} = \T_{\pi_{\theta_{t+1}}} Q_{\eta_t}$ then ${\pi}_{t+2} \propto \pi_{\theta_{t+1}} \exp (\T_{\pi_{\theta_{t+1}}} Q_{\eta_t} - {Q}_{\eta_t}) = \pi_{\theta_{t+1}}$---capturing the advantage of using the proposal $\pi_{\theta_{t+1}}$.

\subsubsection{Optimism in online convex optimization}
One way to design and analyze iterative optimization methods is through online linear optimization (OLO) algorithms.

\paragraph{Online learning}{Policy optimization through the lens of online
learning \citep{Hazan16} means treating the policy optimization algorithm as the learner in online learning and each
intermediate policy that it produces as an online decision. The following steps recast the iterative process of policy optimization into a standard online learning setup: (i) at iteration $t$ the learner plays a decision $\pi_t\in\Pi$, (ii) the environment responds with feedback on the decision $\pi_t$, and the process repeats. The iteration $t$ might be different than the timestep of the environment. Generally, it is assumed that the learner receives an unbiased stochastic approximation as a response, whereas that is not always the case for RL agents, using bootstrapping in their policy gradient estimation with a learned value function.

For an agent it is important to minimize the \textbf{regret} after $T$ iterations
\eqq{
\operatorname{Reg}_T \equiv \sum_{t=0}^{T-1}\left( J(\pi^*) - J(\pi_t)\right)
}
The goal of \textbf{optimistic online learning} algorithms \citep{RakhlinS13, RakhlinS1322, rakhlin2014online} is obtain better performance, and thus guaranteed lower regret, when playing against ``easy'' (i.e., predictable) sequences of online learning problems, where past information can be leveraged to improve on the decision at each iteration.
}

\paragraph{Predictability}{
An important property of the above online learning problems is that they are not completely adversarial. In RL, the policy's true performance objective cannot be truly adversarial,
as the same dynamics and cost functions are used across
different iterations. In an idealized case where the true
dynamics and cost functions are exactly known, using the
policy returned from a model-based RL algorithm would
incur zero regret, since only the interactions with the real
MDP environment, not the model, are considered in the regret minimization problem formulation.
The main idea is to use (imperfect) predictive models, such as off-policy gradients
and simulated gradients, to improve policy learning.
}

\subsubsection{Online learning algorithms}
We now summarize two generalizations of the well-known core algorithms of online optimization for predictable sequences, cf. \citet{joulani20a}: (i) a couple variants of optimistic mirror
descent \citep{chiang12, RakhlinS13, RakhlinS1322, chiang12}, including extragradient descent (\citet{Korpelevich1976TheEM}, and mirror-prox \citep{Nemirovski04, juditsky2011solving}, and (ii) adaptive optimistic follow-the-regularized-leader (AO-FTRL) \citep{RakhlinS13, rakhlin2014online, MohriY16}.

\paragraph{Optimistic mirror descent (OMD). Extragradient methods}{
Starting with some previous iterate $x_t \in \mathcal{X}$, an OMD \citep{joulani20a} learner $x$ uses a prediction $\tilde{g}_{t+1} \in \mathcal{X}^*$ ($\mathcal{X}^*$---dual space of $\mathcal{X}$) to minimize the regret on its convex loss function $f : \mathcal{X} \to \R$ against an optimal comparator $x^* \in \mathcal{X}$ with
\eqq{
x_{t+1} &= \arg\min_{x\in\mathcal{X}} \langle g_{t} + \tilde{g}_{t+1} - \tilde{g}_{t}, x\rangle + \B_{\Omega}(x, x_t)
}
with $\tilde{g}_{t+1} \approx \nabla f(x_{t+1})$ optimistic gradient prediction, and  $g_t \equiv \nabla f(x_t)$ true gradient feedback, $\B_{\Omega}$ a Bregman divergence with mirror map $\Omega$.

Extragradient methods consider two-step update rules for the same objective using an intermediary sequence $\tilde{x}$
\eqq{
\tilde{x}_{t+1} &= \arg\min_{x\in\mathcal{X}} \langle \tilde{g}_{t+1}, x\rangle +\B_\Omega(x, x_{t})
\\
{x}_{t+1} &= \arg\min_{\pi\in\Pi} \langle {g}^+_{t+1}, x\rangle + \B_\Omega(x, x_t)
}
with $\tilde{g}_{t+1} \approx \nabla f(x_{t+1})$ a gradient prediction, and  $g^+_{t+1} \equiv \nabla f(\tilde{x}_{t+1})$ the true gradient direction, but for the intermediary optimistic iterate $\tilde{x}_{t+1}$.
}
\paragraph{Adaptive optimistic follow-the-regularized-leader (AO-FTRL)}{
A learner using AO-FTRL updates $x$ using
\eqq{
x_{t+1} = \arg\min_{x\in\mathcal{X}} \langle g_{0:t} + \tilde{g}_{t+1}, x\rangle +  \omega_{1:t-1}(x)
}
where  $g_{0:t} = \sum_{j=0}^{t} g_j$ are true gradients, $\tilde{g}_{t+1}$ is the optimistic part of the update, a prediction of the gradient before it is received, and $\omega_{0:t}(x) = \sum_{j=0}^{t} \omega_j(x)$ represent the “proximal” part of this adaptive regularization (cf. \citet{joulani20a}), counterparts of the Bregman divergence we have for MD updates that regularizes iterates to maintain proximity.
}
\subsubsection{Policy optimization with online learning algorithms}{
\citet{predictor_corrector} follows the extragradient approach for policy optimization 
\eqq{
\pi_{t+2} &= \arg\max_{\pi\in\Pi} \langle {Q}_{t}, \pi\rangle +\KL(\pi, \pi_{t})
\\
{\pi}_{t+1} &= \arg\max_{\pi\in\Pi} \langle Q_{\pi_t}, \pi\rangle - \KL(\pi, \pi_t)
}
but changes the second sequence to start from the intermediary sequence and add just a correction
\eqq{
\pi_{t+2} &= \arg\max_{\pi\in\Pi} \langle {Q}_{t}, \pi\rangle -\KL(\pi, \pi_{t})
\\
{\pi}_{t+1} &= \arg\max_{\pi\in\Pi} \langle Q_{\pi_t} - {Q}_{t}, \pi\rangle - \KL(\pi, \pi_{t+2})
}
This approach uses  $\pi_{t+2}$ as the optimistic prediction, and ${\pi}_{t+1}$ as the hindsight corrected prediction---a policy optimal in hindsight w.r.t. the average of all previous Q-functions rather than just the most recent one. But it needs an additional model for the value functions $Q_t$, and another learning algorithm to adapt $Q_t$ to the $Q_{t+1}$. Additionally, an agent does not generally have access to $Q_{\pi_t}$, but only partial evaluations.

\citet{Hao2020} also designs an adaptive optimistic algorithm based on AO-FTRL, which updates
\eqq{
\pi_{t+1} = \arg\max_{\pi\in\Pi}\left \langle\left(\textstyle \sum_{j=0}^{t} Q_{j} + \hat{Q}_{t+1}\right), \pi\right\rangle - \alpha_t \omega(\pi)
}
with $ \omega$---a regularizer, and with $Q_{j} \approx Q_{\pi_j}, \forall j \leq t$ predictions for the true gradients, and $\hat{Q}_{t+1} \approx Q_{\pi_{t+1}}$ is also a prediction for the gradient of the next policy, which uses the previous predictions $Q_{\pi_j}, \forall j \leq t$ to compute it. The authors also propose an adaptive method for learning $\alpha_t$ that uses gradient errors of $Q_{j}, \forall j \leq t$.
Averaging value functions has also been explored by \citet{vieillard19} and \citet{vieillard20}.
}

\clearpage
\clearpage
\section{Empirical analysis details}
\label{apend:empirical_analysis}

\begin{table}[h]
\caption{Notation}
\label{table:notation_table_empirical}
\footnotesize
\begin{tabularx}{\textwidth}{p{0.25\textwidth}X}
\toprule
$t$ &  iterations/timesteps
\\
$T$ &  number of iterations
\\
$n$ &  rollout length
\\
$\mathcal{B}$ & buffer
\\
$\mathcal{M}$ & meta-buffer
\\
$w$ & standard critic (Q-fn $Q_w$) parameters
\\
$\eta$ & meta parameters of meta-learner ($u_\eta$ or $U_\eta$)
\\
$\nu$ &  step size for meta-learner's parameters $\eta$ ($U_\eta$)
\\
$\zeta$ &  step size for standard critic's parameters $w$ (Q-fn $Q_w$)
\\
$\xi$ & step size for the policy learner's parameters $\theta$ ($\pi_\theta$) 
\\
$h$ & lookahead horizon
\\
$U$ & search values up to lookahead horizon $h$ (tree depth)
 \\\bottomrule
 \end{tabularx}
\end{table}

\subsection{Algorithms}
\label{apend:algorithms}

\begin{algorithm}[H]
        \caption{\footnotesize\textbf{Policy gradient}}
        \label{alg:pg}
        \begin{algorithmic}[1]
          \STATE \textbf{Init:} params ${\theta_0}$, buffer $\mathcal{B} = [()]$ 
          \FOR {$t \in 0..T$ iterations}
          \STATE  Every $n$ steps using a rollout $\mathcal{B}  \leftarrow (S_t,A_t,R_t,S_{t+1} \dots S_{t+n}) \sim \pi_{\theta_t}$
          \STATE \qquad Update policy learner
          $\pi_{\theta_{t+1}}$ cf. Eq.\ref{eq:pg}
           \ENDFOR
        \end{algorithmic}
      \end{algorithm}

\paragraph{Policy gradients}{
Algorithm~\ref{alg:pg} describes a standard PG algorithm (cf. \citep{Williams92}) with  an expert oracle critic $Q_{\pi_\theta}$, for the policy evaluation of $\pi_\theta$. The standard policy gradient update is 
\eqq{
\theta_{t+1} &= \theta_t + \xi\frac{1}{n}\sum_{i=t}^{t+n} \nabla_{\theta_t} \log \pi_{\theta_t}(A_i|S_i) \left(Q_{\pi_{\theta_t}}(S_i, A_i)  - \E_{\pi_{\theta_t}}[Q_{\pi_{\theta_t}}(S_i, \cdot)]\right)\label{eq:pg}
}
}

\begin{algorithm}[H]
        \caption{\footnotesize\textbf{Actor-critic}}
        \label{alg:ac}
        \begin{algorithmic}[1]
          \STATE \textbf{Init:} params $(\theta_0, w_0)$, buffer $\mathcal{B} = [()]$ 
          \FOR {$t \in 0..T$ iterations}
           \STATE  Every $n$ steps using a rollout $\mathcal{B}  \leftarrow (S_t,A_t,R_t,S_{t+1} \dots S_{t+n})\sim \pi_{\theta_t}$
          \STATE \qquad Update critic
          $Q_{w_{t+1}}$ cf. Eq.\ref{eq:tdq_adv} and  policy learner 
          $\pi_{\theta_{t+1}}$ cf. Eq.\ref{eq:ac}
           \ENDFOR
        \end{algorithmic}
      \end{algorithm}

\paragraph{Actor-critic}{
Algorithm~\ref{alg:ac} describes a standard AC algorithm (cf. \citep{Sutton2000}) with an estimated critic $Q_{w}$, for the policy evaluation of $\pi_\theta$. The policy updates 
\eqq{
\theta_{t+1} &= \theta_t + \xi\frac{1}{n}\sum_{i=t}^{t+n} \nabla_{\theta_t} \log \pi_{\theta_t}(A_i|S_i) \left(Q_{w_t}(S_i, A_i)  - \E_{\pi_{\theta_t}}[Q_{w_t}(S_i, \cdot)]\right)\label{eq:ac}
}
and the critic's update using TD($0$) learning, writes
\eqq{
w_{t+1} &= w_t - \zeta\frac{1}{n}\sum_{i=t}^{t+n}
\left(R_i + \gamma E_{\pi_t}[Q_{w_t}(S_{i+1}, \cdot)] - Q_{w_t}(S_i, A_i)\right)  \nabla_{w_t} Q_{w_t}(S_i,A_i) \label{eq:tdq_adv}
}
}

\begin{algorithm}[H]
        \caption{\footnotesize\textbf{Policy gradients with forward search}}
        \label{alg:forward_search}
        \begin{algorithmic}[1]
          \STATE {\textbf{Init:} params $(\theta_0, w_0)$, buffer $\mathcal{B} = [()]$}
          \FOR {$t \in 0..T$ iterations}
           \STATE  Every $n$ steps using a rollout $\mathcal{B}  \leftarrow (S_t,A_t,R_t,S_{t+1} \dots S_{t+n})\sim\pi_{\theta_t}$
          \STATE \qquad Generate search values ${U}_{{t}}$ up to lookahead horizon $h$ with 
          \STATE \qquad\qquad(i)  ${U}_{{t}} = \T_{\pi_t}^h Q_{w_{t}}$
           \STATE \qquad\qquad (ii) ${U}_{{t}} = \T^h Q_{w_{t}}$
          \STATE \qquad Update critic
          $Q_{w_{t+1}}$ cf. Eq.\ref{eq:td_fw_search} and policy learner 
          $\pi_{\theta_{t+1}}$ cf. Eq.\ref{eq:pg_fw_search}, using ${U}_{{t}}$
           \ENDFOR
        \end{algorithmic}
      \end{algorithm}

\paragraph{Forward search with a model}{
Algorithm~\ref{alg:forward_search} describes an AC algorithm with $h$-step lookahead search in the gradient critic 
\eqq{
\theta_{t+1} &= \theta_t + \xi\frac{1}{n}\sum_{i=0}^n \nabla_{\theta_t} \log \pi_{\theta_t}(A_i|S_i)\left({U}_{t}(S_i, A_i)  - \E_{\pi_t}[{U}_{t}(S_i, \cdot)]\right)\label{eq:pg_fw_search}
}
where ${U}_{t}$ is either (i) ${U}_{t} = \T^h Q_{w_t}$ or (ii) ${U}_{t} = \T^h Q_{w_t}$, depending on the experimental setup, and the critic is updated toward the search Q-values 
\eqq{
w_{t+1} &= w_t - \zeta\frac{1}{n}\sum_{i=0}^n \nabla_{w_t} Q_{w_t}(S_i,A_i) 
\left(R_i + \gamma E_{\pi_t}[{U}_{t}(S_{i+1}, \cdot)] - Q_{w_t}(S_i, A_i)\right) \!\label{eq:td_fw_search}
}
}

\begin{algorithm}[H]
        \caption{\footnotesize\textbf{Optimistic policy gradients with policy targets computed from expert targets}}
        \label{alg:optimistic_pg_expert_tar}
        \begin{algorithmic}[1]
          \STATE \textbf{Init:}  params $(\theta_0, \eta_0)$, buffer $\mathcal{B} = [()]$, meta-buffer $\mathcal{M} = [\mathcal{B}, ..]$ 
           \FOR {$t \in 0..T$ iterations}
            \STATE  Every $n$ steps using a rollout $ \mathcal{B}_t  \leftarrow (S_t,A_t,R_t,S_{t+1} \dots S_{t+n})\sim\pi_{\theta_t}$
            \STATE \qquad Predict ${u}_{\eta_{t-1}} = \varphi(\eta_{t-1}, Q_{\pi_{\theta_{t+1}}})$
          \STATE \qquad Update learner with using optimistic prediction ${u}_{\eta_{t-1}}$ using Eq.~\ref{eq:meta_update}
          \STATE  \qquad Every $h$ steps using experience stored in the meta-buffer $ \mathcal{M} \leftarrow (\mathcal{B}_t, \dots \mathcal{B}_{t+h})$
          \STATE   \qquad  \qquad  Compute policy targets $\pi_{{\theta}_{t+2}}$ cf.  Eq.~\ref{eq:expert_tar_param} or ${\pi}_{t+2}$ cf. Eq.~\ref{eq:expert_tar_geom} 
           \STATE \qquad  \qquad Update meta-learner ${u}_{\eta_{t}}$ cf. Eq.~\ref{eq:opt_pol_grad}
           \ENDFOR
        \end{algorithmic}
      \end{algorithm}

\paragraph{Optimistic policy gradients with expert targets}{
Algorithm~\ref{alg:optimistic_pg_expert_tar} describes a meta-gradient based algorithm for learning \emph{optimistic policy gradients} by supervised learning from policy targets computed with accurate optimistic predictions $Q_{\pi_{t+1}}$.
The meta-update used updates
\eqq{
\theta_{t+1} = \theta_t + \xi u_{\eta_{t-1}} \label{eq:meta_update}
}
where $u_{\eta_t} =\frac{1}{n}\sum_{i=0}^n \nabla_{\theta_t} \log \pi_{\theta_t}(A_i|S_i) 
\left((U_{\eta_t}(S_i, A_i)  - \E_{\pi_{\theta_t}}[(U_{\eta_t}(S_i, A_i)]\right)$
The policy targets are (i) \emph{parametric policies}  obtained at iteration $t$ by starting from the parameters $\theta_{t+1}$ (${\theta}^{0}_{t+2} = \theta_{t+1}$) and executing $h$ parameter updates with data from successive batches of rollouts $\mathcal{B}_{t+1:t+h}$ sampled from the meta-buffer $\mathcal{M}$
\eqq{
{\theta}^{j+1}_{t+2} &= {\theta}^{j}_{t+2}  +\xi \hat{g}^{j}_{t+1}  \label{eq:expert_tar_param}
}
with $\hat{g}^{j}_{t+1} = \frac{1}{n}\sum_{i=0}^n \nabla_{\theta_t} \log \pi_{\theta^{j}_t}(A_i|S_i) 
\left((Q_{\pi_{t+1}}(S_i, A_i)  - \E_{\pi_{\theta_t}}[(Q_{\pi_{t+1}}(S_i, A_i)]\right)$.
After $h$ steps the resulting target parameters ${\theta}_{t+2} \equiv {\theta}^h_{t+2}$, and yield the target policy $\pi_{{\theta}_{t+2}}$. 

The other choice we experiment with is to use a target constructed with (ii) \emph{geometric updates} for one (or more) steps ahead, similarly to tree-search policy improvement procedures. The targets are initialized with $\pi^0_{t+1} = \pi_{\theta_{t+1}}$ and execute one (or more) steps of policy improvement
\eqq{
{\pi}^{j+1}_{t+2} \propto \pi^{j}_{t+2} \exp \alpha {Q}_{\pi_{t+1}} \label{eq:expert_tar_geom}
}
yielding the non-parametric policy target ${\pi}_{t+2} \equiv {\pi}^{j+1}_{t+2}$.
Setting $\alpha\to\infty$ in Eq.\ref{eq:expert_tar_geom}, if the predictions are given, or can be computed with the help of the simulator model, we obtain an update similar to the multi-step greedy operator $\T^h$ used in forward search.

The next parameter vector $\eta_{t}$ for the gradient $u_{\eta}$ is distilled via meta-gradient learning by projecting the expert policy target ${\pi}_{t+2}$ (or $\pi_{{\theta}_{t+2}}$) using the data samples from $\mathcal{B}_t,\dots \mathcal{B}_{t+h}$ from $\mathcal{M}$ and the surrogate objective
\eqq{
\eta_{t+1} &= \eta_t - \nu \frac{1}{h}\sum_{j=t}^{t+h} \nabla_{\eta_{t}} \KL(\pi_{\theta_{t+1}}(S_j), {\pi}_{t+2}(S_j)) \label{eq:opt_pol_grad}
}
}

\begin{algorithm}[H]
        \caption{\footnotesize\textbf{Optimistic policy gradients with target predictions}}
        \label{alg:optimistic_pg_tar_pred}
        \begin{algorithmic}[1]
          \STATE {\textbf{Init:} params $(\theta_0, w_0, \eta_0)$, buffer $\mathcal{B} = [()]$, meta-buffer $\mathcal{M} = [\mathcal{B}, ..]$ }
           \FOR {$t \in 0..T$ iterations}
           \STATE  Every $n$ steps using a rollout $ \mathcal{B}_t  \leftarrow (S_t,A_t,R_t,S_{t+1} \dots S_{t+n})\sim\pi_{\theta_t}$
         \STATE \qquad Predict $u_{\eta_t}$
          \STATE \qquad Update learner with optimistic prediction $u_{\eta_t}$ using Eq.~\ref{eq:meta_update}
          \STATE \qquad Update 
          $Q_{w_{t+1}}$ cf. Eq.\ref{eq:tdq_adv}
          \STATE  \qquad Every $h$ steps using experience stored in the meta-buffer $ \mathcal{M} \leftarrow (\mathcal{B}_t, \dots \mathcal{B}_{t+h})$
           \STATE   \qquad  \qquad  Compute policy targets $\pi_{{\theta}_{t+2}}$ cf.  Eq.~\ref{eq:tar_pred_param} or ${\pi}_{t+2}$ cf. Eq.~\ref{eq:tar_pred_geom} 
          \STATE \qquad  \qquad Update meta-learner $u_{\eta_{t+1}}$ cf. Eq.~\ref{eq:opt_pol_grad}
           \ENDFOR
        \end{algorithmic}
      \end{algorithm}

\paragraph{Optimistic policy gradients with target predictions}
Algorithm~\ref{alg:optimistic_pg_tar_pred} describes a meta-gradient based algorithm for learning \emph{optimistic policy gradients}, by self-supervision from target predictions (learned estimators).
The targets we use are (i) \emph{parametric}, computed at iteration $t$, similarly to the previous paragraph (Eq.~\ref{eq:expert_tar_param}, except we now replace the true optimistic predictions $Q_{\pi_{t+j}}$ with $Q_{w_{t+j}}, \forall j\geq 1$
\eqq{
\hat{g}^{j}_{t+1} = \frac{1}{n}\sum_{i=0}^n \nabla_{\theta_t} \log \pi_{\theta^{j}_t}(A_i|S_i) 
\left((Q_{w_{t+1}}(S_i, A_i)  - \E_{\pi_{\theta_t}}[(Q_{w_{t+1}}(S_i, A_i)]\right)
\label{eq:tar_pred_param}
}
We also experiment with the (ii) \emph{non-parametric target} that takes  \emph{geometric} steps, similarly to Tree-Search policy improvement procedure, where the optimistic prediction from Eq.\ref{eq:expert_tar_geom}, uses the ground truth. 
\eqq{
{\pi}^{j+1}_{t+2} \propto \pi^{j}_{t+2} \exp \alpha {Q}_{w_{t+1}} \label{eq:tar_pred_geom}
}

\subsection{Experimental setup}\label{apend:experimental_setup}
\begin{wrapfigure}{r}{0.26\textwidth} 
\vspace{-40pt}
    \centering
    \includegraphics[width=0.23\textwidth]{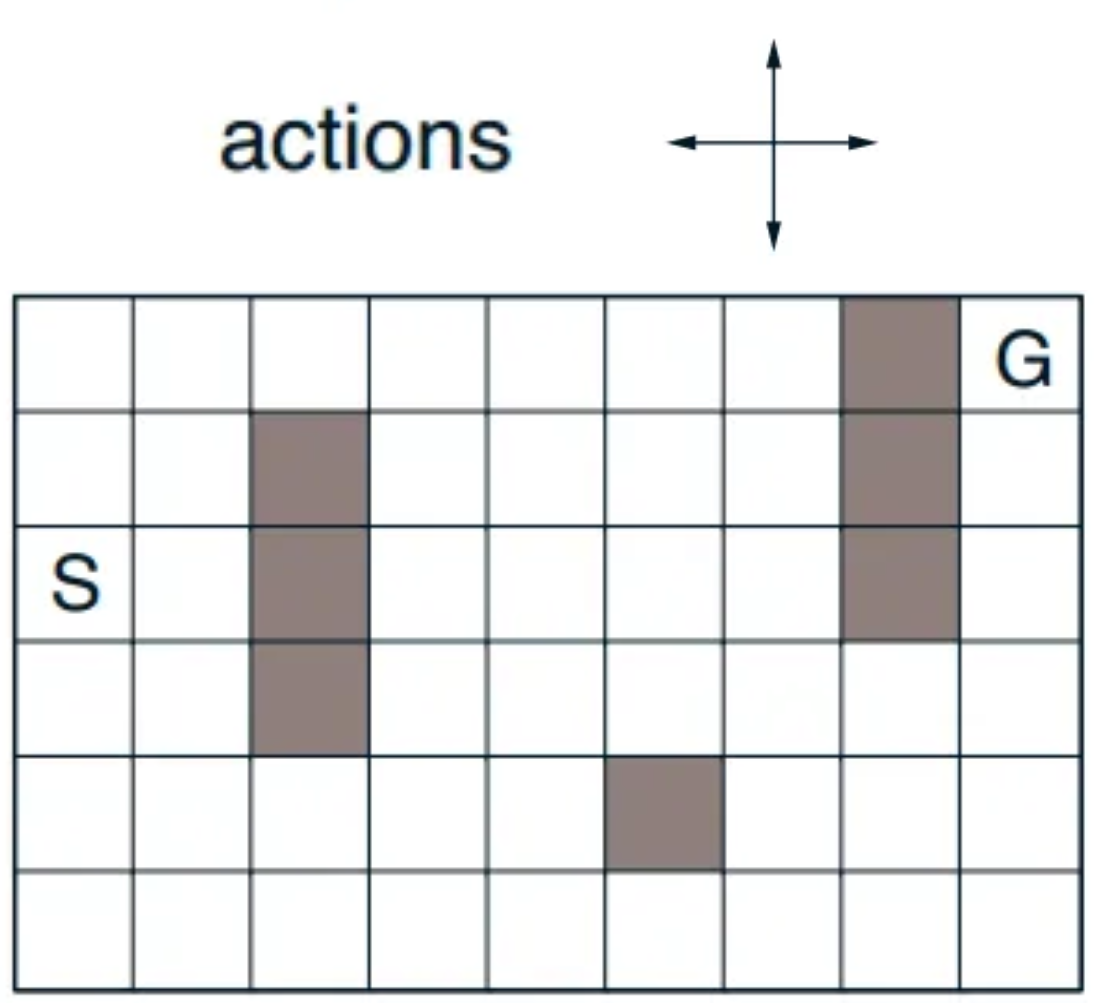}
    \caption{\footnotesize\textbf{Maze Navigation}: illustration of the MDP used in the empirical studies}
    \label{apend:fig:maze}
    \vspace{-20pt}
\end{wrapfigure}

\paragraph{Environment details}
All empirical studies are performed on the same discrete
navigation task from \citet{rl_book}, illustrated in Fig.~\ref{apend:fig:maze}. "G" marks
the position of the goal and the end of an episode. "S" denotes the starting
state to which the agent is reset at the end of the episode. 
The state
space size is $48$, $\gamma = 0.99$. There are $4$ actions that can transition the agent
to each one of the adjacent states. Reward is $1$ at the goal, and zero everywhere else. Episodes terminate and restart from the initial state upon reaching the goal.

\paragraph{Protocol}
All empirical studies report the regret of policy performance every step, and at every episode $J(\pi^*) - J(\pi_t)$, for a maximum number of $500$ episodes. Hyperparameter sensitivity plots show the cumulative regret per total number of steps of experience cumulated in $500$ episodes, $\sum_{t} J(\pi^*) - J(\pi_t)$. This quantity captures the sample efficiency in terms of number of steps of interaction required.

\paragraph{Algorithmic implementation details}
Meta-gradient based algorithms keep parametric representations of the gradient fields via a parametric advantage $\Adv_\eta(s, a) = U_{\eta}(s, a) - E_{\pi}[U_{\eta}(s,A)]$, $\forall s,a$, s.t. a learned gradient update consists of a parametric gradient step on the loss
\eqq{
\theta_{t+1} &= \theta_t + \nabla \mathcal{L}(\theta;\mathcal{B}_t)\Big|_{\theta = \theta_t}
\\
\mathcal{L}(\theta;\mathcal{B}_t) &= \frac{1}{n} \sum _{i=t}^{t+n} \log \pi_\theta(A_i|S_i) \left(U_{\eta}(S_i, A_i) - \sum_a \pi_{\theta_t}(A_i|S_i) U_{\eta}(S_i, A_i)\right)
}
Policies use the standard softmax transform $\pi_{\theta} = \frac{\exp f_{\theta}(s,a)}{\sum_b \exp f_{\theta}(s,b)}$, with $f_{\theta}$, the policy logits. In the experiments illustrated we use a tabular, one-hot representation of the state space as features, so $f_{\theta}$ is essentially  $\theta$. The same holds for the critic's parameter vector $Q_w$, and the meta-learner's parameter vector $U_\eta$.

The experiments were written using JAX, Haiku, and Optax
\citep{bradbury2018jax, deepmind2020jax, haiku2020github}.

\paragraph{Experimental details for the forward search experiment }
We used forward search with the environment true dynamics model up to horizon $h$, backing-up the current value estimate $Q_{w_t}$ at the leaves. We distinguish between two settings: (i) using the previous policy $\pi_t$ for \emph{bootstrapping} in the tree-search back-up procedure, i.e. obtaining ${U}_t = \T^h_{\pi_t} Q_{w_t}$ at the root of the tree; or (ii) using the greedification inside the tree to obtain ${U}_t = \T^h Q_{w_t}$ at the root.
Table~\ref{tab:tp_fw_search} specifies the hyperparameters used for both of the aforementioned experimental settings.   Results shown in the main text are averaged over $10$ seeds and show the standard error over runs. 
\begin{table}[h]
\caption{Hyperparameters for {optimism via forward search} on the Maze Gridworld in Fig.~\ref{apend:fig:maze}}
\label{tab:tp_fw_search}
\footnotesize
\begin{tabularx}{\textwidth}{p{0.25\textwidth}X}
\toprule
\footnotesize
\text{{Hyperparameter}}
&
\text{} 
\\
\hline
\text{$\xi${ (policy step size)}} & 0.5
\\
$\zeta${ (Q-fn step size)}  
 & \{0.01, 0.1, 0.5, 0.9\}
 \\
\text{$h${ (lookahead horizon)}}
 & \{0, 1, 2, 4, 8, 16\}
 \\
\text{$n${ (rollout length)}}
 & 2
 \\
\text{{policy/Q-fn optimiser}}
 & \text{SGD}
\\  \bottomrule
 \end{tabularx}
\end{table}

\paragraph{Experimental details for the meta-gradient experiments with expert target/hints }
For this experiment we used Algorithm~\ref{alg:optimistic_pg_expert_tar} described in Sec.~\ref{apend:algorithms}, and the hyperaparameters in Table~\ref{tab:tp_expert}.

\begin{table}[h]
\caption{Hyperparameters for {optimism via meta-gradient learning with expert targets/hints} on the Maze Gridworld in Fig.~\ref{apend:fig:maze}}
\label{tab:tp_expert}
\footnotesize
\begin{tabularx}{\textwidth}{p{0.25\textwidth}X}
\toprule
\footnotesize
\text{{Hyperparameter}}
\\
\hline
\text{$\xi${ (policy step size)}} & 0.1 
\\
&(\text{training plots Fig.\ref{fig:meta}-d})
\\
&\{0.1, 0.5\} 
\\
&(\text{sensitivity plots Fig.\ref{fig:meta}-c})
\\
$\zeta${ (Q-fn step size)}  
 & - 
 \\
\text{$h${ (lookahead horizon)}}
 & 1 
 \\
 \text{$\alpha${ (step size $\pi$)}}
  & 1 
 \\
\text{$n${ (rollout length)}}
 & 2 
 \\
\text{{policy optimiser}}
 & \text{SGD}  
 \\
 \text{{meta-learner optimiser}}
 & \text{Adam} 
 \\ \bottomrule
 \end{tabularx}
\end{table}

\paragraph{Experimental details for the meta-gradient experiments with target predictions/bootstrapping }
For this experiment we used Algorithm~\ref{alg:optimistic_pg_tar_pred} described in Sec.~\ref{apend:algorithms}, and the hyperaparameters in Table~\ref{tab:tp_pred}.
\begin{table}[h]
\caption{Hyperparameters for {optimism via meta-gradient learning with target predictions/bootstrapping} on the Maze Gridworld in Fig.~\ref{apend:fig:maze}}
\label{tab:tp_pred}
\footnotesize
\begin{tabularx}{\textwidth}{p{0.25\textwidth}X}
\toprule
\footnotesize
\text{{Hyperparameter}}
\\
\hline
\text{$\xi${ (policy step size)}}  & 0.5 
\\
$\zeta$ { (Q-fn step size)}  
& 0.1 
\\
&(\text{training plots Fig.\ref{fig:meta}-e})
\\
&\{0.1, 0.5\} 
\\
&(\text{sensitivity plots Fig.\ref{fig:meta}-f})
 \\
\text{$h${ (lookahead horizon)}}
 & 1 
 \\
\text{$n${ (rollout length)}}
 & 2 
 \\
\text{{policy optimiser}}
 & \text{SGD}  
 \\
 \text{{meta-learner optimiser}}
 & \text{Adam} 
 \\ \bottomrule
 \end{tabularx}
\end{table}
\clearpage

\subsection{Additional results \& observations}{
 Fig~\ref{apend:fig:1meta_pg} shows learning curves, and Fig~\ref{apend:fig:2meta_pg} hyperparameter sensitivity, for  experiments with expert targets, when using Adam for the meta-optimization.  Fig~\ref{apend:fig:3meta_pg} (learning curves), and  Fig~\ref{apend:fig:4meta_pg} (hyperparameter sensitivity) illustrate results for when the meta-optimization uses SGD. The next set of figures show experiments with target predictions---for Adam
 Fig~\ref{apend:fig:1meta_ac} (learning curves) and Fig~\ref{apend:fig:2meta_ac} (hyperparameter sensitivity), and for SGD  Fig~\ref{apend:fig:3meta_ac} (learning curves) and Fig~\ref{apend:fig:4meta_ac} (hyperparameter sensitivity).

\begin{figure}
\vspace{-10pt}
\footnotesize
\centering
 \subfigure[\footnotesize expert targets, Adam, perf/episode]{\label{apend:fig:meta_pg:a}\includegraphics[width=0.35\textwidth]{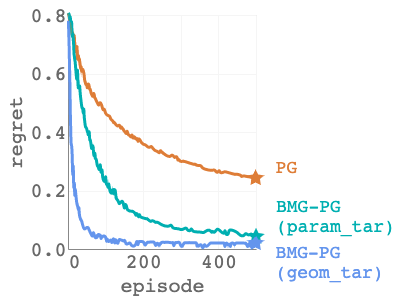}}
 \hspace{50pt}
 \subfigure[\footnotesize expert targets, Adam, perf/step]{\label{apend:fig:meta_pg:b}\includegraphics[width=0.35\textwidth]{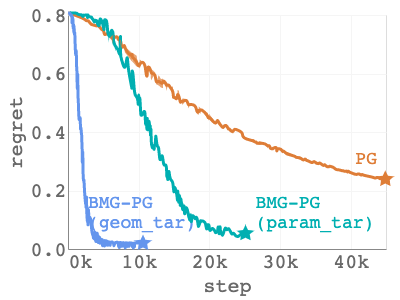}}
\caption{\footnotesize
 Meta-learner uses Adam. Policy optimization with adaptive optimistic policy gradients. x-axis - (a) no of episodes, (b) no of steps.  y-axis - regret $J(\pi^*) - J(\pi_t)$. Learning curves denote: the baseline - {\color{retro_orange} standard PG} algorithm, \emph{adaptive optimistic policy gradient learning algorithms} - with {\color{sea_green} parametric target policies} , {\color{cornflower_blue} functional non-parametric target policies}, trained with meta-gradients from \emph{expert targets}.
 Shades (wherever noticeable) denote standard error over different runs.}
     \label{apend:fig:1meta_pg}
\end{figure}

\begin{figure}
\vspace{-10pt}
\centering
  \subfigure[\footnotesize Hyperparam. sensitivity / expert targets / Adam / total regret]{\label{apend:fig:meta_pg:c}\includegraphics[width=0.35\textwidth]{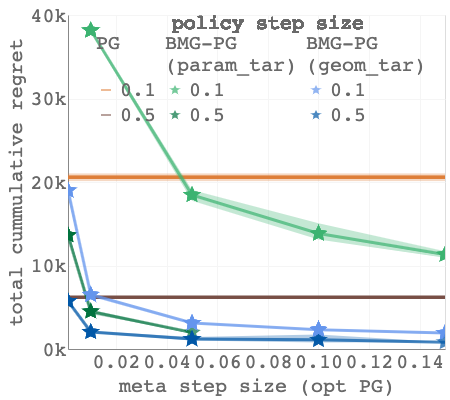}}
     \hspace{50pt}
    \subfigure[\footnotesize Hyperparam. sensitivity / expert targets / Adam / final regret]{\label{apend:fig:meta_pg:d}\includegraphics[width=0.35\textwidth]{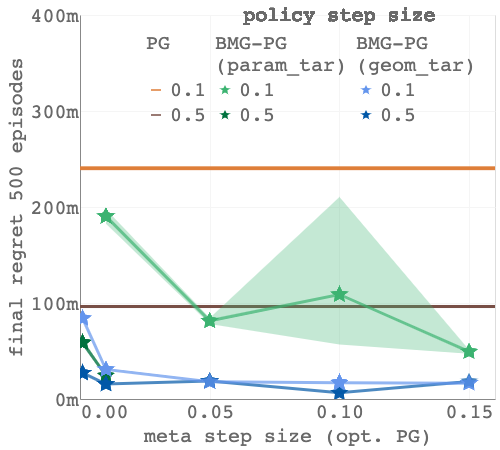}}
\caption{\footnotesize
 Meta-learner uses Adam.
Hyper-parameter sensitivity curves for the meta-learning rate $\nu$ - x-axis, y-axis - total cummulative regret (a): $ \sum_{i\leq t} J(\pi^*) - J(\pi_i)$, final regret (b): $  J(\pi^*) - J(\pi_T)$; Learning curves show \emph{adaptive optimistic policy gradient learning algorithms} - with {\color{sea_green} parametric target policies} , {\color{cornflower_blue} functional non-parametric target policies}, trained with meta-gradients from \emph{expert targets}.
Different tones show the evolution of the meta-hyperparameter $\nu$ to those used in the inner learned optimization algorithm, i.e. the policy step size.
Different straight lines denote the baseline---{\color{retro_orange} standard PG}. Shades denote standard error over different runs.}
     \label{apend:fig:2meta_pg}
\end{figure}

\begin{figure}
\vspace{-10pt}
\footnotesize
\centering
    \subfigure[\footnotesize expert targets, SGD, perf/episode]{\label{apend:fig:meta_pg:e}\includegraphics[width=0.35\textwidth]{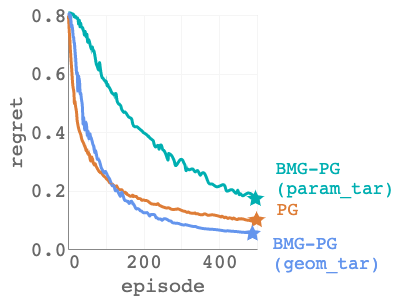}}
     \hspace{50pt}
 \subfigure[\footnotesize  expert targets, SGD, perf/step]{\label{apend:fig:meta_pg:f}\includegraphics[width=0.35\textwidth]{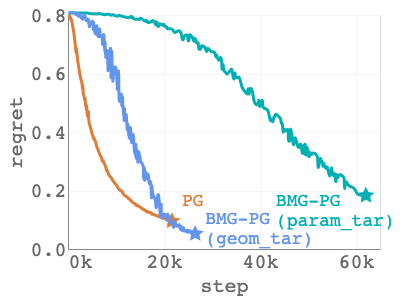}}
\caption{\footnotesize Meta-learner uses SGD. Policy optimization with adaptive optimistic policy gradients. x-axis - (a) no of episodes, (b) no of steps. y-axis - regret $J(\pi^*) - J(\pi_t)$.  Learning curves denote: the baseline - {\color{retro_orange} standard PG} algorithm, \emph{optimistic policy gradient learning algorithms} - with {\color{sea_green} parametric target policies} , {\color{cornflower_blue} functional non-parametric target policies}, trained with meta-gradients from \emph{expert targets}. Shades (wherever noticeable) denote standard error over different runs.}
     \label{apend:fig:3meta_pg}
\end{figure}

\begin{figure}
\vspace{-10pt}
\centering
  \subfigure[\footnotesize expert targets, SGD, total regret]{\label{apend:fig:meta_pg:g}\includegraphics[width=0.35\textwidth]{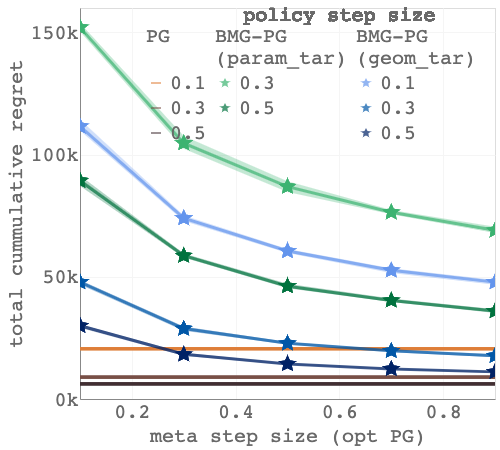}}
         \hspace{50pt}
    \subfigure[\footnotesize expert targets, SGD, final regret]{\label{apend:fig:meta_pg:h}\includegraphics[width=0.35\textwidth]{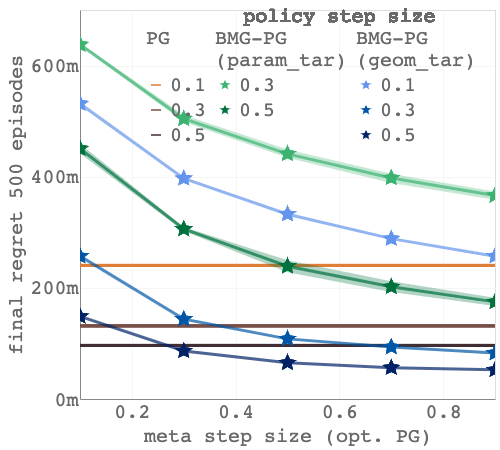}}
\caption{\footnotesize
Meta-learner uses SGD. Hyper-parameter sensitivity curves for the meta-learning rate $\nu$ - x-axis. y-axis - total cumulative regret (a): $ \sum_{i\leq t} J(\pi^*) - J(\pi_i)$, final regret (b): $  J(\pi^*) - J(\pi_T)$. Learning curves show \emph{adaptive optimistic policy gradient learning algorithms} - with {\color{sea_green} parametric target policies}, {\color{cornflower_blue} functional non-parametric target policies}, trained with meta-gradients from \emph{expert targets}.
Different tones show the evolution of the meta-hyperparameter $\nu$ to those used in the inner learned optimization algorithm, i.e. the policy step size.
Different straight lines denote the baseline---{\color{retro_orange} standard PG}. Shades denote standard error over different runs.}
     \label{apend:fig:4meta_pg}
\end{figure}

\begin{figure}
\vspace{-10pt}
\centering
 \subfigure[\footnotesize  target pred, Adam, perf/episode]{\label{apend:fig:meta_ac:a}\includegraphics[width=0.35\textwidth]{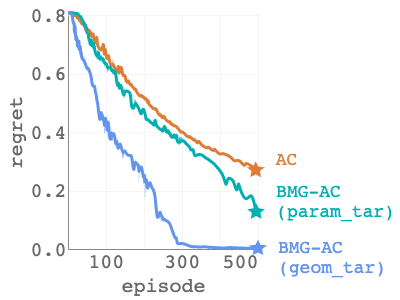}}
        \hspace{50pt}
 \subfigure[\footnotesize target pred, Adam, perf/step]{\label{apend:fig:meta_ac:b}\includegraphics[width=0.35\textwidth]{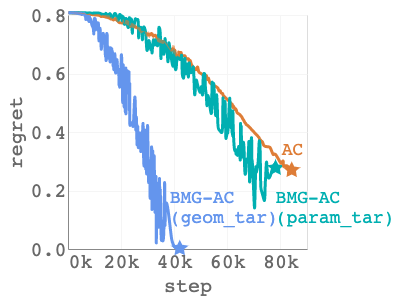}}
\caption{\footnotesize
 Meta-learner uses Adam. Policy optimization with adaptive optimistic policy gradients. x-axis - (a) no of episodes, (b) no of steps. y-axis - regret $J(\pi^*) - J(\pi_t)$. Learning curves denote: the baseline - {\color{retro_orange} standard AC} algorithm, \emph{optimistic policy gradient learning algorithms} - with {\color{sea_green} parametric target policies} , {\color{cornflower_blue} functional non-parametric target policies}, trained with meta-gradients from \emph{target predictions}. Shades (wherever noticeable) denote standard error over different runs.}
     \label{apend:fig:1meta_ac}
\end{figure}

\begin{figure}
\vspace{-10pt}
\centering
  \subfigure[\footnotesize target pred, Adam, total regret]{\label{apend:fig:meta_ac:c}\includegraphics[width=0.35\textwidth]{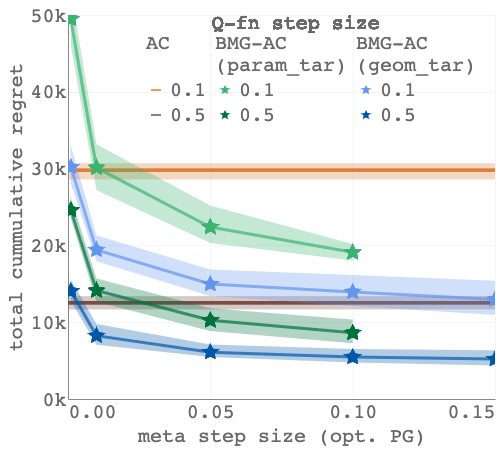}}
       \hspace{50pt}
    \subfigure[\footnotesize target pred, Adam, final regret]{\label{apend:fig:meta_ac:d}\includegraphics[width=0.35\textwidth]{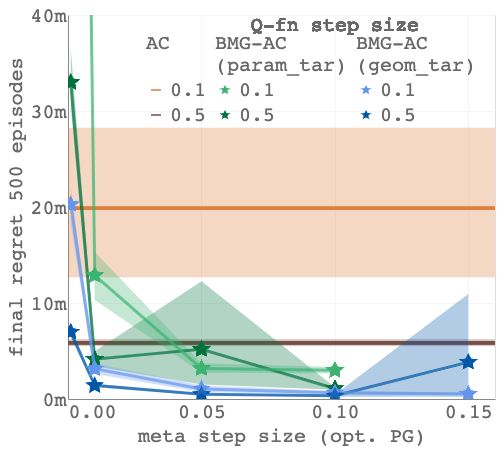}}
\caption{\footnotesize
Meta-learner uses Adam. Hyper-parameter sensitivity curves for the meta-learning rate $\nu$ - x-axis. y-axis - total cummulative regret (a): $ \sum_{i\leq t} J(\pi^*) - J(\pi_i)$, final regret (b): $  J(\pi^*) - J(\pi_T)$. Learning curves show \emph{optimistic policy gradient learning algorithms} - with {\color{sea_green} parametric target policies} , {\color{cornflower_blue} functional non-parametric target policies}, trained with meta-gradients from \emph{target predictions}. 
Different tones show the evolution of the meta-hyperparameter $\nu$ to those used in the inner learned optimization algorithm, i.e. Q-fn step size. 
Different straight lines denote the baseline---{\color{retro_orange} standard AC}. Shades denote standard error over different runs.}
     \label{apend:fig:2meta_ac}
\end{figure}

\begin{figure}
\vspace{-10pt}
\centering
    \subfigure[\footnotesize target pred, SGD, perf/episode]{\label{apend:fig:meta_ac:e}\includegraphics[width=0.35\textwidth]{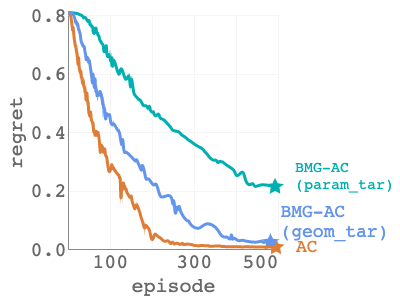}}
        \hspace{50pt}
 \subfigure[\footnotesize target pred, SGD, perf/step]{\label{apend:fig:meta_ac:f}\includegraphics[width=0.35\textwidth]{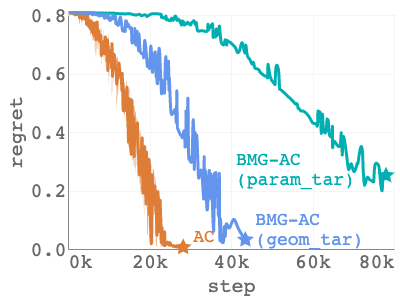}}
\caption{\footnotesize
 Meta-learner uses SGD. Policy optimization with optimistic policy gradients. x-axis - (a) no of episodes, (b) no of steps. y-axis - regret $J(\pi^*) - J(\pi_t)$.  Learning curves denote: the baseline - {\color{retro_orange} standard AC} algorithm, \emph{optimistic policy gradient learning algorithms} - with {\color{sea_green} parametric target policies}, {\color{cornflower_blue} functional non-parametric target policies}, trained with meta-gradients from \emph{target predictions}. 
 Shades (wherever noticeable) denote standard error over different runs.}
     \label{apend:fig:3meta_ac}
\end{figure}

\begin{figure}
\vspace{-20pt}
\centering
  \subfigure[\footnotesize target pred, SGD, total regret]{\label{apend:fig:meta_ac:g}\includegraphics[width=0.35\textwidth]{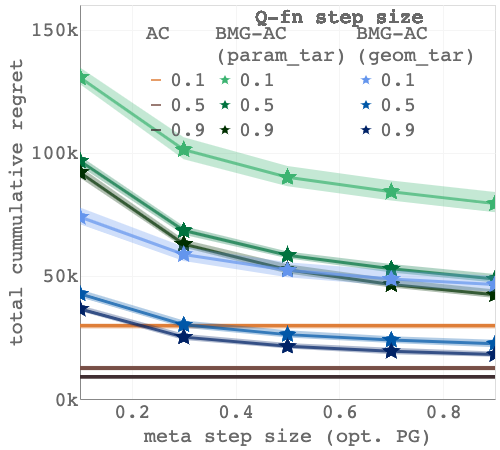}}
       \hspace{50pt}
    \subfigure[\footnotesize target pred, SGD, final regret]{\label{apend:fig:meta_ac:h}\includegraphics[width=0.35\textwidth]{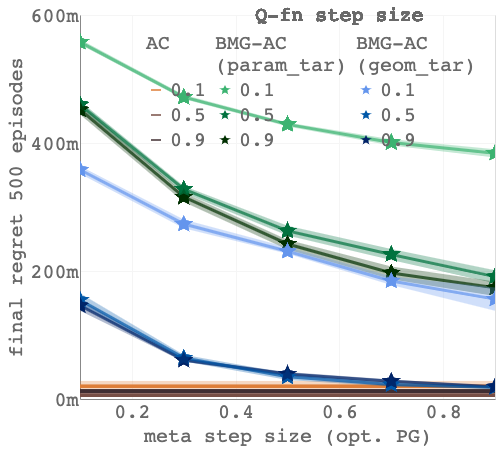}}
\caption{\footnotesize Meta-learner uses SGD.
Hyper-parameter sensitivity curves for the meta-learning rate $\nu$ - x-axis; y-axis - total cummulative regret (a): $ \sum_{i\leq t} J(\pi^*) - J(\pi_i)$, final regret (b): $  J(\pi^*) - J(\pi_T)$. Learning curves show \emph{optimistic policy gradient learning algorithms} - with {\color{sea_green} parametric target policies} , {\color{cornflower_blue} functional non-parametric target policies}, trained with meta-gradients from \emph{target predictions}. 
Different tones show the evolution of the meta-hyperparameter $\nu$ to those used in the inner learned optimization algorithm, i.e. Q-fn step size. 
Different straight lines denote the baseline---{\color{retro_orange} standard AC}. Shades denote standard error over different runs.}
     \label{apend:fig:4meta_ac}
\end{figure}

}

\clearpage

\clearpage
	\small{
	\bibliographystyle{abbrvnat}
		\bibliography{bib}
		}

\end{document}